\documentclass[journal]{IEEEtran}

\usepackage{rotating}

\usepackage{booktabs}

\usepackage{enumitem}
\usepackage{multirow}

\usepackage{epsfig}

\usepackage{amsmath}
\usepackage{amssymb}

\usepackage{graphicx}

\usepackage{subcaption}
\usepackage{footmisc}
\usepackage[most]{tcolorbox}

\DeclareCaptionLabelSeparator{periodspace}{.\quad}
\usepackage{hyperref}

\ifCLASSINFOpdf
\else
\fi

\begin{document}

\title{Benchmarking the Robustness of Instance Segmentation Models}

\author{ Yusuf Dalva$^{*}$, Hamza Pehlivan$^{*}$\thanks{*Joint first authors,  contributed equally.}, Said Fahri Altındiş and Aysegul Dundar
\thanks{Y. Dalva is with Virginia Tech, USA, email: ydalva@vt.edu}
\thanks{H. Pehlivan is with MPI, Germany.}
\thanks{S. F. Altindis, and A. Dundar are with the Department of Computer Science, Bilkent  University,
Ankara, Turkey.} 
\thanks{e-mail: \{ fahri.altindis\}@bilkent.edu.tr,\\ adundar@cs.bilkent.edu.tr )}
\thanks{Corresponding author is A. Dundar.}
}

\maketitle

\begin{abstract}
This paper presents a comprehensive evaluation of instance segmentation models with respect to real-world image corruptions as well as out-of-domain image collections, e.g. images captured by a different set-up than the training dataset. 
The out-of-domain image evaluation shows the generalization capability of models, an essential aspect of real-world applications and an extensively studied topic of domain adaptation.
These presented robustness and generalization evaluations are important when designing instance segmentation models for real-world applications and picking an off-the-shelf pretrained model to  directly use for the task at hand.
Specifically, this benchmark study includes state-of-the-art network architectures, network backbones, normalization layers, models trained starting from scratch versus pretrained networks, and the effect of multi-task training on robustness and generalization.
Through this study, we gain several insights. For example, we find that group normalization enhances the robustness of networks across corruptions where the image contents stay the same but corruptions are added on top. On the other hand, batch normalization improves the generalization of the models across different datasets where statistics of image features change.
We also find that single-stage detectors do not generalize well to larger image resolutions than their training size.
On the other hand, multi-stage detectors can easily be used on images of different sizes.
We hope that our comprehensive study will motivate the development of more robust and reliable instance segmentation models.

\end{abstract}

\begin{IEEEkeywords}
Robustness, Image Corruptions, Domain Adaptation, Instance Segmentation, Deep Networks.
\end{IEEEkeywords}

\IEEEpeerreviewmaketitle

\section{Introduction}

Convolutional Neural Networks (CNNs) have achieved impressive results on various computer vision tasks for scene understanding \cite{dundar2016embedded, ren2015faster, liu2016ssd,  chen2018encoder, dundar2021unsupervised, li2021survey, eryilmaz2022understanding}. These CNN models are evaluated and continuously improved based on hyper-parameter and architectural searches conducted on clean validation subsets. Such validation subsets share similarities with their corresponding training sets, e.g., collected with the same camera, at a similar time of the day and weather \cite{cordts2016cityscapes}, and sometimes even in the same room \cite{ionescu2013human3}. 
However, when these models are deployed for real-world applications, they may be used to infer on various domains, such as from images with corruptions to images collected by different cameras and with different pre-processings applied.
Therefore, it is vital to evaluate models across different domains to measure their robustness, generalization, and reliability. 

Understanding and analyzing the robustness of deep networks for corruption is an important topic since corrupted images naturally exist in many forms.
Corruption in the image may be caused by environmental factors such as occlusions on the camera lens due to  rain, mud, or frost; the image blurs due to a fast-moving camera, and various noise corruptions due to a defect in the recording hardware or sensor. 
Hence, it is important that we build robust models across image corruptions.
Additionally, when a model is deployed, test images, even when they are clean, may exhibit significant differences with the training sets the model was trained on.
For example, they may be recorded with different aspect ratios, different camera parameters, and under different illuminations.
These differences in the test and train time data cause considerable degradation in terms of accuracy, and many works propose to adapt models onto new test domains, which is an important research topic known as domain adaptation \cite{tzeng2014deep,long2015learning,sun2016deep,dundar2020domain}.
Domain adaptation methods require a pipeline for each test domain and may be impractical if there is no fixed test domain.
Therefore, building robust models that generalize well across domains is another important research endeavor. 

In this work, we conduct an extensive evaluation of instance segmentation models across various real-world domains.
We do not target adapting a CNN model to a pre-defined domain such as a specific type of corruption. This would require specific actions for that domain, e.g. adding their corresponding augmentations. Such an assumption would be impractical given the large number of existing domains.
Instead, we are interested in evaluating models trained on clean datasets with respect to a broad range of image corruptions and images collected from different environments that exhibit domain gaps with the training data.

We are interested in instance segmentation models because of their wide range of applications.
They are an important part of autonomous driving applications \cite{cordts2016cityscapes} and robotics \cite{xie2021unseen}. 
Additionally, off-the-shelf instance segmentation models are used to generate labels to train other tasks such as image synthesis \cite{dundar2020panoptic, yildirim2023inst} as well as 3D reconstruction \cite{bhattad2021view, dundar2022fine}.
Considering the application domains, these models are expected to achieve good accuracies on a diverse set of images.
For a task at hand, while it may be attractive to pick the best performing model on the clean validation set to deploy, it may not be the model that generalizes best to other domains.
In this paper, we aim to investigate the instance segmentation models' robustness across domains and find the ingredients of the models that enhance the robustness.

In summary, our contributions can be summarized as follows:
\begin{itemize}
    \item Comprehensive benchmarking of instance segmentation algorithms with different network architectures, network backbones, normalization layers, and network initializations.
    \item Measuring robustness across a wide range of real-world image corruptions as well as measuring robustness across out-of-domain image collections, which is an important evaluation set-up for domain adaptation.
    \item Measuring the effect of joint-task training on robustness by comparing detection and instance segmentation models.
    \item Based on this study, we provide several new insights. We find that normalization layers play an important role in robustness.
    Single-stage instance segmentation models are more robust to image corruptions. On the other hand, multi-stage instance segmentation models achieve better generalizations to other image collections that contain objects with a wide range of scales.
    Finally, we find that network backbones and copy-paste augmentations affect robustness significantly.
    Surprisingly, ImageNet pre-training compared to training from scratch does neither affect the robustness nor the generalization of the model. We provide more observations and insights in the Results section.
\end{itemize}

\section{Related Work}
\label{sec:related}

In this section, we first review related work on robustness studies and domain adaptations.
Then we extensively review instance segmentation methods that are included in this paper.

\subsection{Robustness studies} 
Recently, deep networks are evaluated under real-world image corruptions for image classification \cite{geirhos2018generalisation, hendrycks2019benchmarking}, object detection \cite{michaelis2020benchmarking}, and segmentation \cite{kamann2020benchmarking} tasks and found to have degraded performance in terms of accuracy compared to when tested on clean images.
Heavy occlusions \cite{ke2021deep} are the other challenges to the robustness of instance segmentation models, but we leave it out of the scope for this work.
Various methods are proposed to improve the robustness of deep networks \cite{zheng2016improving, jin2015robust}.
One popular approach is to increase their
network bias towards object shapes which has become popular after the discovery that deep networks trained on ImageNet are biased towards texture \cite{geirhos2018imagenet}.
Researchers propose to provide shape bias to networks with data augmentations, for example, by stylizing training images \cite{geirhos2018imagenet, michaelis2020benchmarking}, and by alpha-blending, a portion of RGB training images with texture-free representations, which are semantic maps with randomly chosen colors that are fixed for each semantic class \cite{kamann2020increasing}.
Domain randomization is another technique to provide a shape bias by aggressively augmenting synthetic data with various color manipulations  \cite{tobin2017domain}. It is shown to be successful in generalizing a model learned from synthetic data to work on real-world cases \cite{tobin2017domain, tremblay2018training}.

\subsection{Domain adaptation} Studies have demonstrated that deep learning methods suffer from significant degradation in terms of accuracy when the test time input feature distributions differ from the training datasets \cite{tzeng2014deep,long2015learning}.
The change in the distribution may come from different noise conditions \cite{saenko2010adapting}, synthetic versus real image datasets  \cite{dundar2020domain}, among images that are collected in different cities, with different lighting, and season \cite{yu2020bdd100k} just to name a few.
Since the differences in distributions may also come from real-world image corruptions, improving robustness and generalizing the model to an unseen domain go hand in hand. 
The adaptation is usually proposed to adapt a machine learning model to a pre-defined target dataset by aligning two domains of images via adversarial training ~\cite{taigman2016unsupervised,shrivastava2016learning}, aligning feature levels in networks ~\cite{ganin2015unsupervised,hoffman2016fcns,tzeng2017adversarial}, and network output level~\cite{tsai2018adapt}, or a combination of several \cite{dundar2020domain}.
In this paper, we are not interested in adapting a network to a pre-defined target dataset but in finding out which recipes show improved robustness across unseen domains.

\subsection{Instance Segmentation Studies}
Instance segmentation is a challenging computer vision task that requires correct pixel-level and instance-level predictions simultaneously.
There are dedicated surveys on topics of object detection and instance segmentation \cite{zhao2019object, jiao2021new, liu2023survey} for further interested readers.
In this section, we review different instance segmentation models and recipes that are included in our benchmark.

\subsubsection{Network Architectures}

In this benchmarking study, we investigate different choices of instance segmentation architectures. 
We categorize them as multi-stage and single-stage architectures.

\noindent \paragraph{Multi-Stage Architecture}
 They include Mask-RCNN  \cite{he2017mask},  PointRend \cite{kirillov2020pointrend}, QueryInst \cite{fang2021instances} frameworks. Mask-RCNN models are one of the most popular instance segmentation models with many variations that include dilated convolution, feature pyramid network, and versions with deformable convolution and cascaded architectures. 
 These multi-stage architectures are built on Faster-RCNN and follow a detect and then segment paradigm.
 They are explained below:
 
\begin{itemize}[leftmargin=*]
    \item \textbf{C4}  - The original implementation of Faster R-CNN \cite{ren2015faster} with ResNet component extracting features from the final convolutional layer of the 4th stage, which is referred to as C4. R50-C4 denotes this backbone with ResNet-50 architecture.
    \item \textbf{DC5} - Dilated convolution \cite{yu2016multiscale} is a  modified version of the default convolution algorithm where calculations of some points are skipped according to the given set of dilation parameters.
    As a consequence, it has a larger receptive field than the standard convolution layers. In \cite{dai2017deformable}, dilation is used on the fifth convolutional layer of ResNet architecture. This method is referred as DC5.
    \item \textbf{FPN} - Feature Pyramid Network  \cite{lin2017feature} is a feature extractor that can generate feature maps at different scales from a single-scale image in a fully convolutional manner.
    FPN includes two main components: bottom-up and top-down pathways. The bottom-up pathway is the usual convolutional network for feature extraction as it calculates feature maps at different scales. The computed outcome of each stage's last residual block is used as feature activations for ResNets. On the other hand, the top-down pathway creates higher resolution features by upsampling feature maps from higher pyramid levels.
    The generated features are improved with features from the bottom-up pathway using lateral connections to obtain these improved features.
    \item \textbf{FPN-Deform.} - Deformable convolution augments the spatial sampling locations in the modules with additional offsets providing better flexibility compared to the default convolutional layer, which is limited to geometric model transformations due to the fixed geometric structures in their building modules \cite{dai2017deformable, zhu2019deformable}. Deformable convolution can adapt to the geometric variations of objects and has shown to be beneficial in object detection, instance segmentation tasks \cite{dai2017deformable, zhu2019deformable} and has been successfully incorporated for image synthesis \cite{mardani2020neural,altindis2023refining}.
    Deformable convolution is used in the FPN model by replacing the last set of  convolution layers.
    \item \textbf{FPN-Cas.} - Cascade R-CNN \cite{cai2018cascade, cai2019cascade} is a multi-stage extension of the R-CNN, where the detector stages deeper into the cascade, and the model becomes sequentially more selective against close false positives.
    The cascaded architecture is extended to Mask-RCNN with the addition of a mask head to the last stage.
    The cascaded head is added to the FPN architecture and referred to as FPN-Cas.
    
    \item \textbf{PointRend} -  Point-based Rendering \cite{kirillov2020pointrend} module is inspired by the sampling methods used for efficient rendering algorithms. PointRend starts with a coarse instance segmentation prediction and adaptively samples points on the image in a higher resolution to achieve fine-detailed instance predictions.
    It replaces the default mask head of Mask-RCNN with a module that iteratively computes predictions of a set of points with a small MLP (multi-layer perceptron).
    
    \item \textbf{QueryInst} - 
   Recently, query-based set predictions are used to tackle object detection and instance segmentation tasks \cite{carion2020end, dong2021solq, fang2021instances}. 
    Instances as queries \cite{fang2021instances} is one of them, which uses dynamic mask heads in the query-based end-to-end detection framework.
    The dynamic mask heads are set in parallel with each other, and they transform each mask RoI feature adaptively according to the corresponding query.
    They are simultaneously trained in all stages. 
   During inference,  all the dynamic mask heads in the intermediate stages are thrown away, and only the final stage predictions are used for inference.
     
    \end{itemize}

\noindent \paragraph{Single-Stage Architecture} They include TensorMask \cite{chen2019tensormask},
SOLO \cite{wang2020solo} (SOLO, SOLOv2), YOLACT \cite{bolya2019yolact} (YOLACT, YOLACT++), and PolarMask \cite{xie2020polarmask}.
Single-stage architectures are more challenging to build and first successful single-stage methods are developed after second-stage architectures were well-established such as Mask-RCNNs.

\begin{itemize}[leftmargin=*]

 \item \textbf{TensorMask} \cite{chen2019tensormask} is a single-stage model that  uses a dense sliding window approach to produce instance masks.
    The instance predictions are encoded with 4D tensors for each window.
    This method achieves almost comparable results with Mask-RCNN.
   
    \item \textbf{SOLO}  \cite{wang2020solo} is another one-stage instance segmentation model which estimates masks based on their locations without outputting any bounding boxes.
    An image is divided into a grid, and according to the coordinates of the object center, an object instance is assigned to one of the grid cells.
    There is an instance mask prediction for each grid, and it is outputted as the final prediction if a semantic category is detected for that grid.
    Since the predictions of instance masks need to be spatially variant,  normalized pixel coordinates are fed to a convolutional layer at the beginning of the network. 
    ResNet backbones and FPN architecture are used by default.
    
    \item \textbf{SOLOv2} \cite{wang2020solov2} improves upon SOLO by replacing the mask predictions that are calculated for each grid separately by decoupled mask kernel prediction and mask feature learning.
    Mask kernel predictions are used as convolution weights on the mask features.
    This dynamic instance segmentation model provides a more efficient mask representation and learning and additionally enables higher resolution mask predictions for finer details.

    \item \textbf{PolarMask} \cite{xie2020polarmask} predicts the contour of instance through instance center classification and dense distance regression in a polar coordinate via fully convolutional single-stage architecture.
    It uses FPN to extract features of different levels.

    \item \textbf{YOLACT} \cite{bolya2019yolact} is an another one stage instance segmentation model.
    The architecture is similar to YOLO \cite{redmon2016you}, a single-stage detector. 
    YOLACT extends the bounding box detection branch to estimate mask coefficients in addition to the bounding box locations and class confidence.
    The mask coefficients are used to linearly combine the prototypes, which are estimated in parallel via a second branch.
    The architecture is built on various backbones with FPN.
    
    \item \textbf{YOLACT++} \cite{bolya2020yolact++} improves on YOLACT by utilizing deformable convolution layers, optimized numbers of anchors, and a mask re-scoring regime \cite{huang2019mask}. Deformable convolutions are used in the last 3 ResNet stages with an interval of 3 (i.e., skipping two ResNet blocks in between), resulting in total 11 deformable layers.
  
 \end{itemize}

\subsubsection{Network Backbones}
\label{sec:net_bone}
We experiment with  instance segmentation models based on different types of backbone architectures
that are both revolutionary in terms of network design and comprehensive in terms of instance segmentation scores. In this regard, we experiment using ResNet \cite{he2016deep}, ResNeXT \cite{xie2017aggregated}, SpineNet \cite{du2020spinenet}, Swin \cite{liu2021swin}, and MViTV2-T \cite{li2022mvitv2}.
\begin{itemize}[leftmargin=*]
    \item \textbf{ResNet} - 
    ResNet architecture \cite{he2016deep} is commonly used for various computer vision tasks due to its accuracy. For our robustness analysis, we use ResNet-50 (R50) and ResNet-101 (R101) architectures, where the number indicates the depth of the backbone networks. 
    These networks involve residual blocks formed by three consecutive convolutional blocks, as specified by the original study.
    In these backbones, the residual connection is defined as an element-wise summation of extracted feature maps. 
    \item \textbf{ResNeXt} -  In addition to the existence of residual connections between building blocks, ResNeXt \cite{xie2017aggregated} introduces the concept of \textit{split-transform-merge} to backbones with residual connections. Originating from the representation of a neuron combining input features with learned weights, ResNeXt offers a network structure aggregating set of transformations.
    This structure introduces cardinality to building blocks, which denotes the number of transformation blocks to be aggregated.
    With these stronger representations, the ResNeXt  structure achieves learning a more complex representation compared to ResNet variants.
    In order to assess the robustness of this increase in predictive performance, we include a ResNeXt containing 101 layers in our experiments and refer to it as X101.
    
    \item \textbf{SpineNet} -  This architectural paradigm adopts scale-permuted architecture design instead of scale-decreased methods \cite{du2020spinenet}.
    The study introducing this methodology argues that the success of architectures that apply feature down-scaling in consecutive levels does not translate well to tasks requiring simultaneous recognition and localization.
    Different than scale-decreased methods, SpineNet backbones are constructed after performing a \textit{Neural Architecture Search (NAS)} and consider permutations of different design decisions on network design in a learnable way using reinforcement techniques.
    We include SpineNet-49/96/143 architectures in our robustness studies, where the numerical value defines the number of layers.

    \item \textbf{Swin} - Transformer based architectures have been recently popularized and achieve state-of-the-arts on many computer vision tasks \cite{dosovitskiy2020image}. Transformer-based backbones split images into patches and treat each patch as a word embedding. Patches are processed individually, followed by an attention mechanism across patches. We experiment with a recently proposed Swin architecture \cite{liu2021swin}, which computes cross-windows self-attention by shifting windows between layers so that attention computation is not limited to the same patches.
    We experiment with two variants of Swin architecture, SwinT, and SwinS, which have a complexity similar to those of ResNet-50 and ResNet101, respectively. 
    
    \item \textbf{MViTV2-T} - Another transformer based architecture we experiment with is MViTV2-T \cite{li2022mvitv2}, Multiscale Vision Transformers.
   MViTV2-T proposes to  process high-resolution visual input by using pooling attention to overcome computational cost. 
   Experiments of MViTV2-T show that their proposed pooling attention  is more effective than local window attention, which is used by Swin architectures.

\end{itemize}

\newcommand{\interpfigi}[1]{\includegraphics[width=4.2cm,height=2.4cm]{#1}}
\begin{figure*}[]
  \centering
  \setlength\tabcolsep{0.32pt}
  \renewcommand{\arraystretch}{0.15}
  \begin{tabular}{cccc}
    Clean & Motion B. & Defocus B. & Impulse N. \\
    \interpfigi{}&
    \interpfigi{}&
    \interpfigi{}&
    \interpfigi{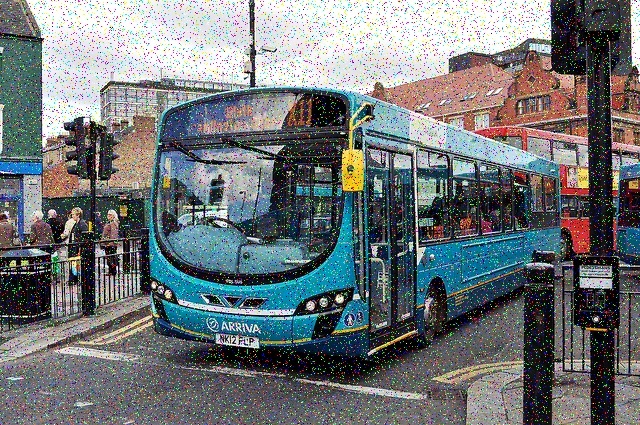}\\
    \interpfigi{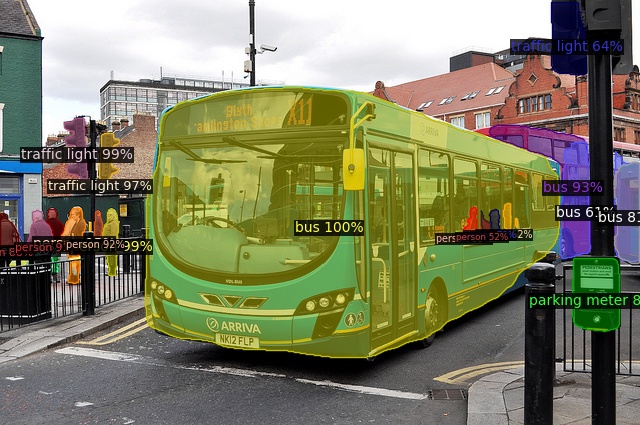}&
    \interpfigi{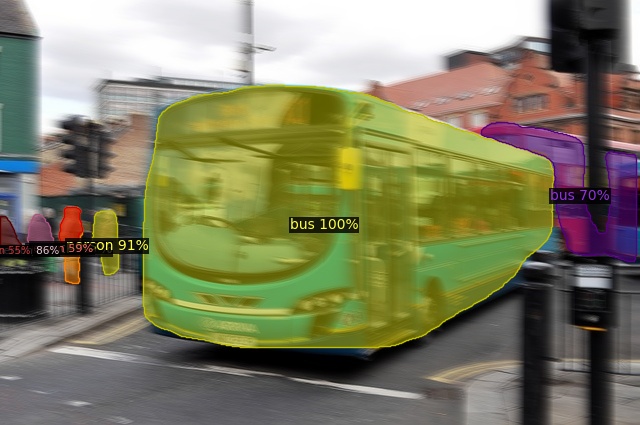}&
    \interpfigi{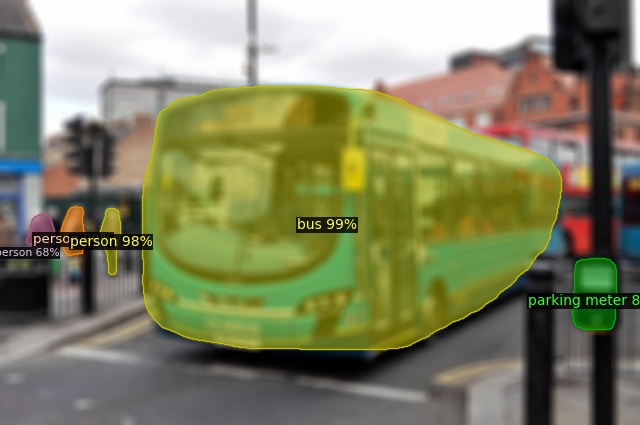}&
    \interpfigi{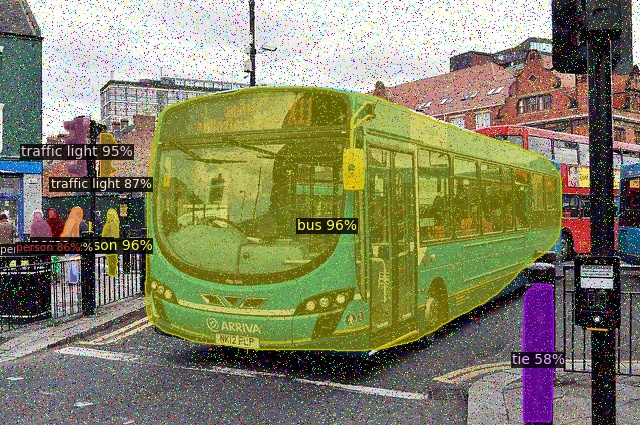}\\
    Brightness D.  & JPEG. D. & Snow W.&  Frost W.\\
    \interpfigi{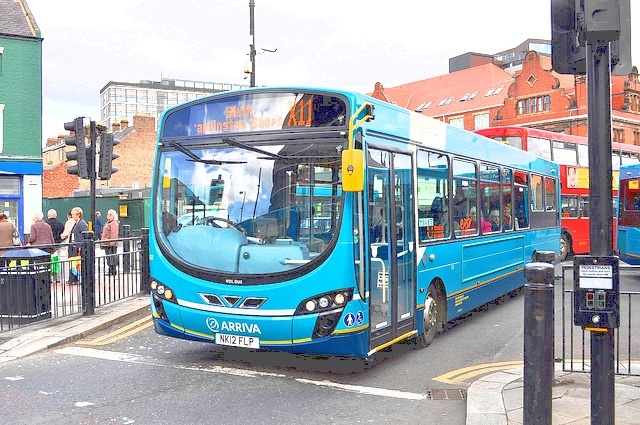}&
    \interpfigi{}&
    \interpfigi{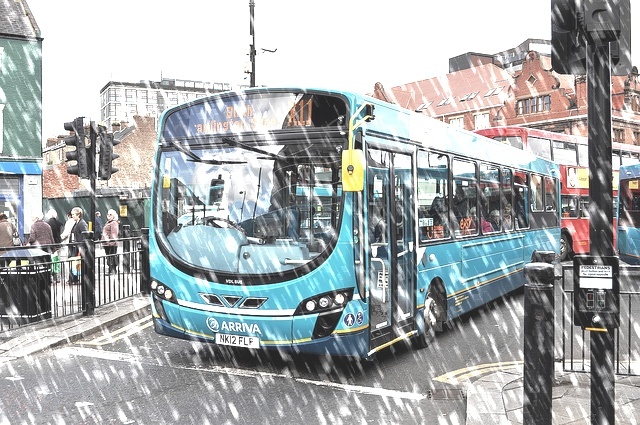}&
     \interpfigi{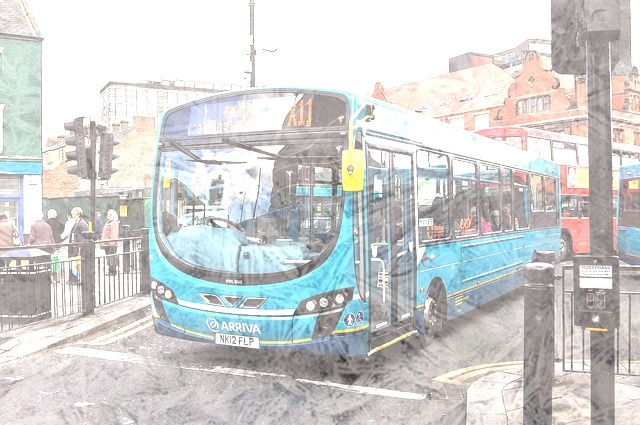}\\
    \interpfigi{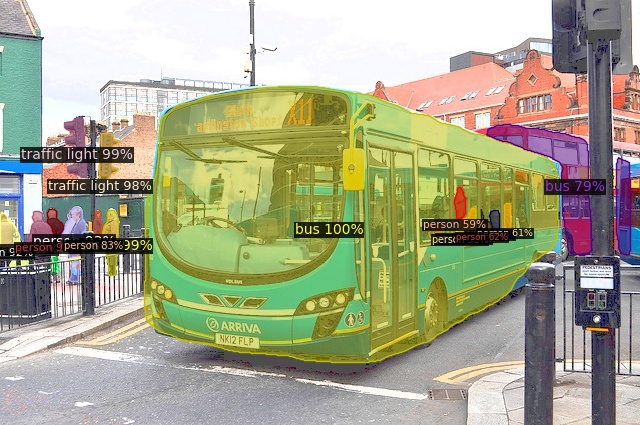}&
    \interpfigi{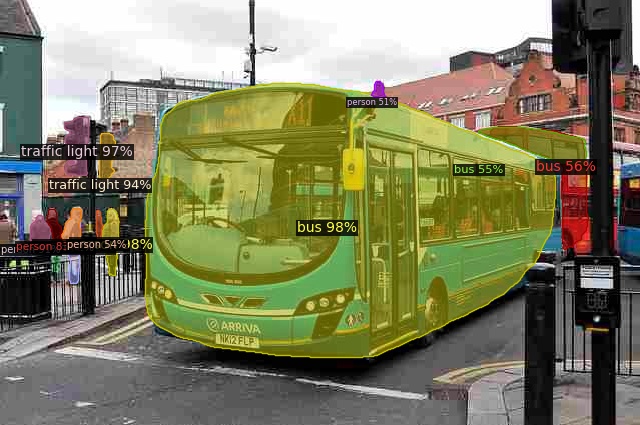}&
    \interpfigi{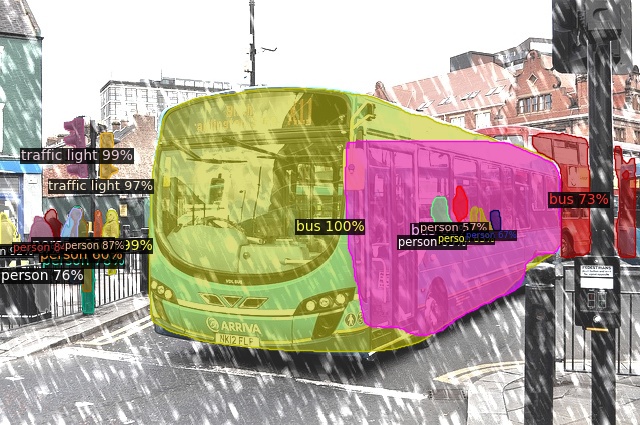}&
    \interpfigi{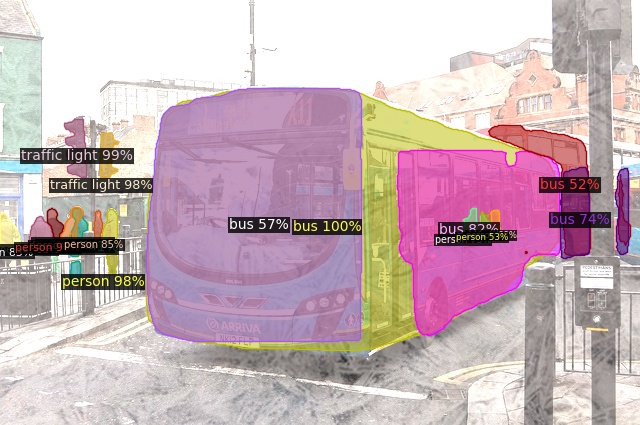}\\
     \end{tabular}
     \caption{An image from the COCO validation dataset, the original clean version, and corrupted ones by different noise models (severity=3). B.=Blur, N.=Noise, D.=Digital, and  W.=Weather. Row 2 and 4 show inference results of the R50-FPN model on images from rows 1 and 3, respectively.
     On blur corrupted images, traffic lights and most of the persons are no longer detected.
     With impulse noise, the pole is detected as a tie. While changing the brightness of an image does not cause a significant accuracy drop, the parking meter is no longer detected in this example.
     The model does not detect persons on the bus and falsely detects a person on top of the bus on the image with JPEG compression.
     Snow and frost cause falsely duplicated detections on the bus, which are accurately detected on the other corrupted images.}
     \label{fig:corruption_examples}
\end{figure*}

\subsubsection{ImageNet Pretraining and Normalization Layers}

Since the initial breakthrough of deep learning applied to object detection and instance segmentation, fine-tuning networks that were pretrained on the ImageNet dataset has been a popular recipe \cite{ren2015faster,  liu2016ssd,  lin2017feature, he2017mask}. 
However, it is recently shown that ImageNet pre-training may not be beneficial if the dataset is big enough \cite{he2019rethinking, zoph2020rethinking}. 
Furthermore, they show that better accuracies are obtained when starting from random initialization at the expense of longer iterations on the COCO dataset.
Our experiments investigate if models trained from random initialization show less or more robustness to image corruptions than their ImageNet pre-trained counterparts.

We couple these experiments with different choices of normalization layers.
Normalization layers, since the discovery of batch normalization  \cite{ioffe2015batch} layer, have become an important ingredient of deep neural networks.
Various normalization layers are proposed for different tasks \cite{ioffe2015batch, ulyanov2016instance, ba2016layer,wu2018group}
that are applied to different groupings of input features.
Since they collect statistics of input features from datasets, their role in robustness is significant as image corruptions or out-of-domain images may exhibit different feature statistics. 
Among these normalization layers, we compare the models that use batch normalization, synchronized batch normalization, and group normalization layers since they are shown to work well on instance segmentation models.

\begin{itemize}[leftmargin=*]

\item Batch Normalization (BN) \cite{ioffe2015batch} is a feature normalization module that is found to be helpful to train deeper and more accurate networks. 
BN normalizes the features by computing the mean and standard deviation for each channel in each batch. BN, as in other normalization methods, learns a trainable scale and shift parameter for each channel.
Since the statistics are collected over a batch, small batch sizes severely degrade the accuracy of BN.
When pretrained networks are tuned for instance segmentation tasks, the batch statistics of the pretrained model can be frozen since instance segmentation models can only fit small batches to a GPU, given that they are trained on high-resolution images.
Otherwise, BN cannot be used to train instance segmentation models from scratch.

\item Synchronized Batch Normalization (SyncBN) \cite{liu2018path} is an extension of BN that collects batch statistics computed across multiple devices (GPUs).
SyncBN is an essential module for training models that can only fit a few images to each GPU, such as instance segmentation model trainings.
When few images, e.g. only 1-2 images, fit a GPU, BN cannot compute robust statistics, and SyncBN resolves this issue by increasing the effective batch size when using many GPUs.
This extension also enables training from scratch instance segmentation models.
 
 \item  Group Normalization (GN) \cite{wu2018group} is one of the proposed alternatives to BN. 
Different than other feature normalization methods, GN divides channels into groups and computes statistics per group. 
Therefore, with this grouping, the computation is independent of the batch dimension as it operates on each example independently.

\end{itemize}

\subsubsection{Copy-Paste Augmentation}
Data augmentation is an essential part of computer vision applications to increase the effective data size and improve results.
Other than standard augmentation techniques like color and scale jittering, datasets with instance annotations allow  copy-paste augmentation of instances.
Copy-paste augmentation is recently performed with a simple pipeline where between two randomly sampled images,  instances from one image are pasted onto the other after both images going through scale augmentations, creating a large variety of unique training images \cite{ghiasi2021simple}.
With this approach, significant improvements are obtained in instance segmentation scores with longer trainings.
For example, where training a network from scratch without such augmentation requires around 100 epochs to converge (the validation set performance no longer improves), the models trained with this augmentation require around 400 epochs. Even though the improvements over the baseline are already visible around epoch 100, longer training further boosts the accuracy.
This benchmarking study investigates the robustness of these models, which are trained with and without copy-paste augmentation. 
We experiment with this augmentation because it is specifically proposed for the instance segmentation task and achieve significant improvements.

\section{Methods}

We evaluate the robustness of instance segmentation models under a wide range of image corruptions \cite{hendrycks2019benchmarking}.
The corruptions are applied to validation sets and not used during training.
Models are trained on non-corrupted (clean) COCO training set \cite{lin2014microsoft}.
Additionally, we investigate robustness across image collections that exhibit domain gap with COCO. For this setting, we use models trained on the COCO training set and validate them on the Cityscapes \cite{cordts2016cityscapes} and BDD \cite{yu2020bdd100k} datasets.

\subsection{ImageNet-C Corruptions}
We investigate many image corruptions from the ImageNet-C dataset \cite{hendrycks2019benchmarking}.
These corruptions are mainly categorized into four categories, several types of  \textbf{1) Blur} (motion, defocus, gaussian), \textbf{2) Noise} (gaussian, impulse, shot and speckle), \textbf{3) Digital} (brightness, contrast, and JPEG compression), and \textbf{4) Weather} (snow, spatter, fog, and frost). 
Examples of these corrupted images are shown in Fig. \ref{fig:corruption_examples}. Each corruption can be tuned for different severity. We find severity set to 3 looks realistic and challenging enough and use this setting as the standard in our evaluations.
Corruptions are as follows:

\textbf{1.} Blur corruptions include \textit{motion blur}, which results from a fast-moving camera when recording.
\textit{Defocus blur} happens when an image is out of focus.
\textit{Gaussian blur} is the last blur corruption, which is implemented with a low-pass filter and outputs a blurred pixel by weighted averaging a pixel with its neighbors.
The libraries and parameters used for these corruptions are given in Table \ref{tab:method_blur}.

\begin{table}[h]
\caption{Blur corruption parameters.}\label{tab:method_blur}
\centering
\renewcommand\tabcolsep{4pt}
\begin{tabular}{p{3.2cm}|p{2.2cm}|p{1.8cm}}
\toprule
 Motion & Defocus & Gaussian   \\
 \hline
Wand motion blur library with radius 15, sigma 8, and a randomly sampled angle from a uniform distribution between (-45, 45). & 
OpenCV Gaussian blur function with $3\times3$ kernels, and sigmaX values of 0.5. & 
The skimage gaussian filters are used with a sigma value of 3. \\
\hline
 \end{tabular}
 \end{table}

\textbf{2.} Among noise corruptions, \textit{Gaussian noise} appears in low-lighting conditions.
It is implemented as a summation of the input image and a noise map where the noise is generated from a normal distribution with a scale of $0.18$.
\textit{Impulse noise} naturally happens due to a defect in recording hardware or sensor and corrupts some of the pixels in an image. It is implemented as a color analogue of salt-and-pepper noise by using the skimage random noise function with an amount of $0.09$.
\textit{Shot noise} is electronic noise caused by the discrete nature of light itself. 
It is implemented with a Poisson distribution parameterized by the input image multiplied by $12$.
The image is then normalized by $12$ to be in the correct range.
\textit{Speckle noise} occurs due to the effect of environmental conditions on the imaging sensor during image acquisition. It is an additive noise and the noise added to a pixel tends to be proportional to the original pixel intensity. 
We use the parameter of $0.35$ for the normal distribution's scale and multiply it with the input image to output the noise, which is then added to the input image.

\textbf{3.} Digital corruptions include \textit{brightness}, \textit{contrast}, and \textit{saturation}.
These types actually may not be considered as corruptions, since they are natural characteristics of images with their intensity varies with daylight, lighting conditions, and the photographed object’s color.
They are also commonly manipulated in image editing tools to make images more colorful and bright.
For brightness, images are converted to HSV from RGB, and $0.3$ is added to the last channel and converted back to RGB.
For contrast, the mean of the image is removed, and the image is multiplied by $0.2$ before the mean is added back.
For saturation, images are again converted to HSV, and the first channel is multiplied by $2$.
Additionally, we look at the corruption caused by JPEG format, a commonly used lossy image compression format, and introduce compression artifacts.
We save the images with JPEG quality of $15$.

\textbf{4.} Weather corruptions include \textit{snow}, \textit{spatter}, \textit{frost}, and \textit{fog}. Among these, spatter and frost corruptions are modeling the effect of these weathers on the camera lenses.
Spatter can occlude a lens in the form of rain or mud, and frost forms when lenses are coated with ice crystals.
Implementations of these methods include many parameters, and we use the severity 3 for the given packages \cite{hendrycks2019benchmarking}.

\begin{figure}[]
  \centering
  \setlength\tabcolsep{0.32pt}
  \renewcommand{\arraystretch}{0.15}
  \begin{tabular}{cc}
     Clean & Mixed Corruption (MC)  \\
    \interpfigi{} &
    \interpfigi{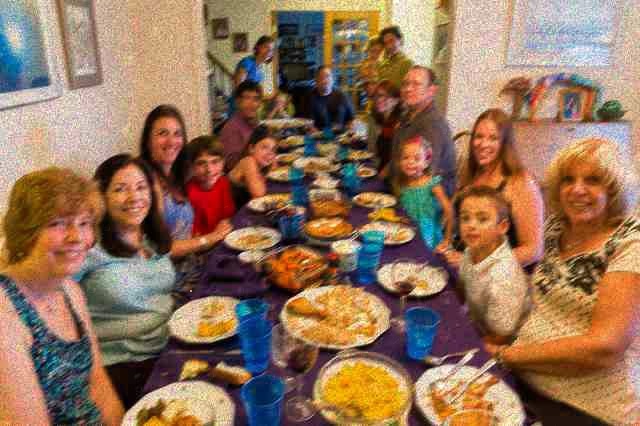}\\
    \interpfigi{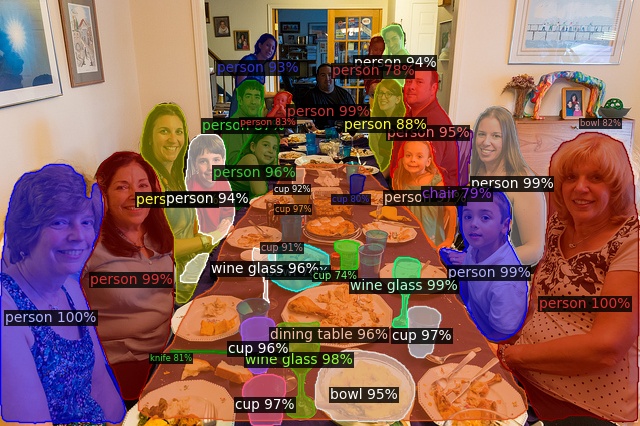}&
    \interpfigi{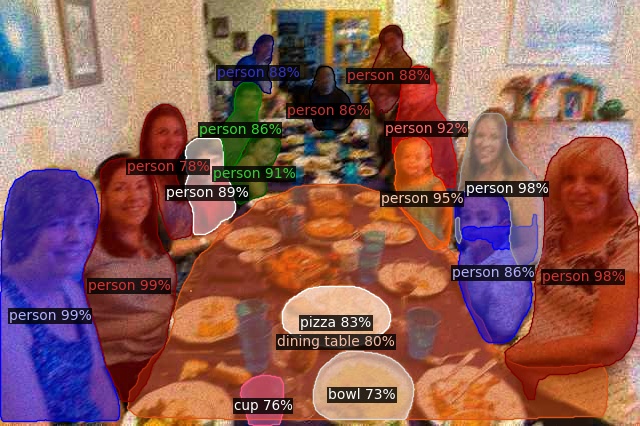}\\
     \end{tabular}
     \caption{An image from the COCO validation dataset, the original clean version, and the corrupted one under mixed corruption setting. This image is corrupted with motion blur, speckle noise, jpeg compression, and spatter.
     We show the inference results of R50-FPN model in Row 2.}
     \label{fig:corruption_mixed}
\end{figure}

\begin{figure}[]
  \centering
  \setlength\tabcolsep{0.32pt}
  \renewcommand{\arraystretch}{0.15}
  \begin{tabular}{cc}
     Clean & Geometric Transform. (GTr)  \\
    \interpfigi{} &
    \interpfigi{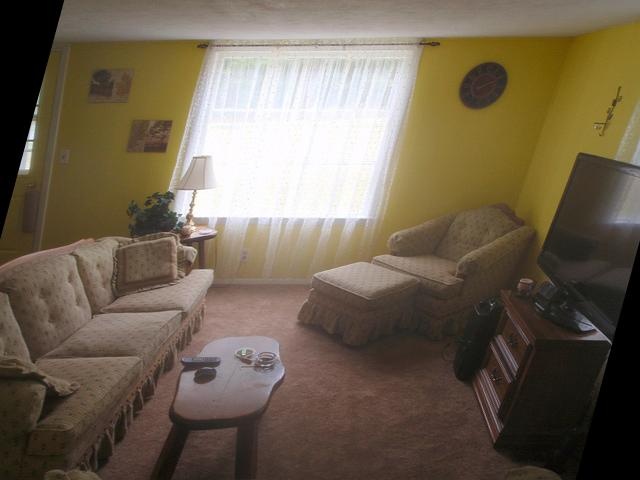}\\
    \interpfigi{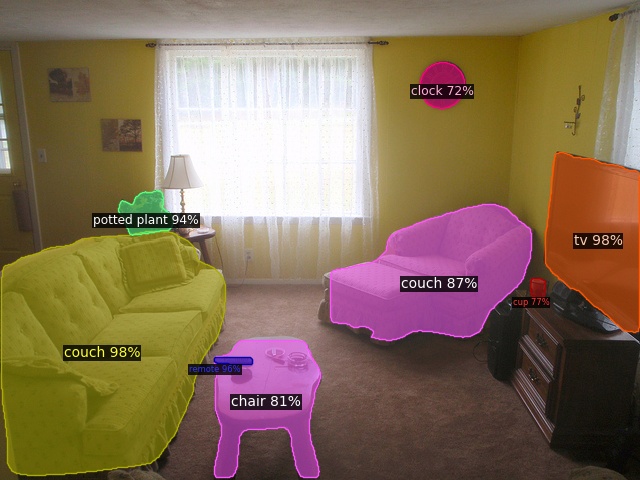}&
    \interpfigi{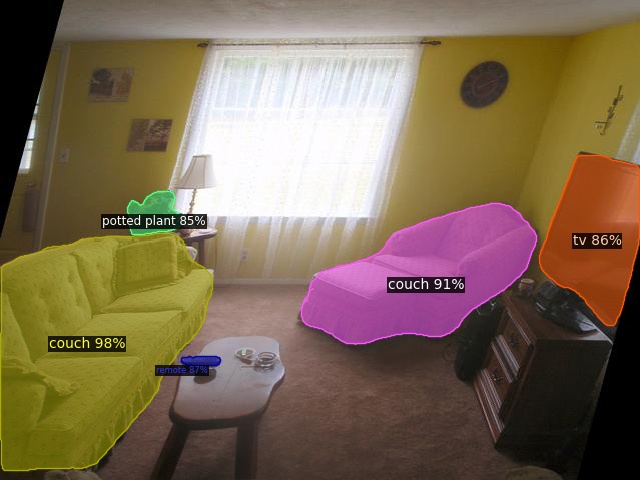}\\
     \end{tabular}
     \caption{An image from the COCO validation dataset, the original clean version, and the corrupted one under geometric transformation. 
     We show the inference results of R50-FPN model in Row 2.}
     \label{fig:corruption_affine}
\end{figure}

\subsection{Mixed Corruptions}

We evaluate the models under mixed corruptions by corrupting each image with 4 corruptions from the main categories, blur, noise, digital, and weather. We randomly pick a type for each of these corruption categories. For example, from the blur category, we pick one of the motion, defocus, and gaussian corruption types.
We set the severity to 1 in this setting since 4 different corruptions are added to each image. When severity is equal to 3, images are not recognizable, therefore, not providing a reasonable setting. Severity parameters can be found from the corruption library\footnote{https://github.com/hendrycks/robustness/tree/master/ImageNet-C}. An example of the mixed corruption dataset images is shown in Fig. \ref{fig:corruption_mixed}.

\subsection{Geometric Transformations}
We also run experiments by applying realistic geometric transformations as shown  in Fig. \ref{fig:corruption_affine} to further test instance segmentation model's robustness on different view points.
We apply shear transformation\footnote{https://imgaug.readthedocs.io/en/latest/source/overview/geometric.html} on images and transform the polygons that represent masks for each instance accordingly.
We set the shear degree to $15$ to both achieve realistic and challenging scenes as shown in  Fig. \ref{fig:corruption_affine}.
We refer to this corruption as \textit{GTr.} in Tables.
 
\subsection{Cross Dataset}
Cross-dataset evaluation, training a network on a source dataset and evaluating it on a different, target dataset, is a standard setup to evaluate domain adaptation methods.
In such a setup, methods additionally use unlabeled images from the target dataset.
However, we are not interested in adopting a method with an unlabeled target dataset but in finding the model that achieves the best score on the unseen target dataset without any re-training.
For this, we use the Cityscapes \cite{cordts2016cityscapes} and BDD datasets \cite{yu2020bdd100k} as our target datasets.
Cityscapes consist of urban street scene images from various cities in Germany.
This dataset is collected for autonomous driving applications with cameras mounted on a car as it drives through and therefore shows significant differences (shown in Fig, \ref{fig:citys_noise}) with COCO dataset \cite{lin2014microsoft}, which is our source dataset. 
This dataset has annotations of eight instance classes: person, rider, car, truck, bus, train, motorcycle, and bicycle. Among these instance classes, COCO does not include rider.
Given that riders are detected as persons by COCO labeling guidelines, 
we relabel a rider as a person in the ground truth and run the evaluation on seven classes.
We evaluate the models both on the Cityscapes validation set (500 images) and training dataset (2975 images).
The second dataset we use is the BDD dataset which includes images from New York, San Francisco, and Bay Area with diverse scene types such as city streets, residential areas, and highways.
Some diverse examples are shown in Fig. \ref{fig:bdd_noise}.
BDD has the same classes as the Cityscapes, and we follow the same setup for evaluation.
We use the BDD validation dataset, which has 1000 images.
We infer on Cityscape and BDD images on their original resolutions, $1024\times2048$ and $720\times1280$, respectively.
The cross-dataset experiments also measure the robustness across geometric distortions such as varying scales, objects viewed from varying angles, scale transformations, and occlusions.

\subsection{Joint-task Training}
We also investigate if joint-task training helps to improve the robustness of deep networks.
We compare an object detection model and instance segmentation model in which addition to bounding boxes, instance masks are estimated.
We compare these models with respect to their bounding box detection AP.
Previously, it is shown that learning instance masks over bounding box detection provides better performance for detection showing benefits of joint-task training \cite{he2017mask}.
In this paper, we study if such behavior exists for  corrupted and out-of-domain image collections.
The results of this setup are important as they may lead practitioners to add additional tasks into their training to increase robustness.

\subsection{Evaluation Metrics}
We report AP metric averaged over IoU (intersection over union) thresholds.
In instance segmentation experiments, we report the AP evaluation using mask IoU. For joint-task experiments, we report the AP evaluation using bounding boxes.
We calculate APs for each corrupted dataset.
We additionally report the average of them (APs on each corrupted dataset) as given in Eq. \ref{eqn:map}.
\begin{equation}
    \label{eqn:map}
    AP_{avg} = \frac{1}{C} \sum_{i=1}^C AP_i
\end{equation}
where $i$ corresponds to various corruptions. We exclude the AP calculation on the clean validation dataset from this averaging.
With this metric, we can observe which models are more robust to the set of corruptions mentioned before.
We additionally report Corruption Degradation (CD) and relative Corruption Degradation (rCD) following other robustness works \cite{hendrycks2019benchmarking, kamann2020benchmarking} as given in Eq. \ref{eqn:cd} and Eq. \ref{eqn:rcd}, respectively.

\begin{equation}
    \label{eqn:cd}
    CD^f = \frac{1}{C} \sum_{i=1}^C \frac{D_i^f}{D_i^{ref}}
\end{equation}

\begin{equation}
    \label{eqn:rcd}
    rCD^f = \frac{1}{C} \sum_{i=1}^C \frac{D_i^f-D_{clean}^f}{D_i^{ref}-D_{clean}^{ref}}
\end{equation}

The degradation in this context is defined as $D=1-AP$, $i$ corresponds to different corruptions, and $f$ corresponds to different models.
We consider the degradation of models relative to clean data by subtracting $D_{clean}$ from $D_i$. Additionally, we divide the corruption degradation by a reference model, $ref$.
With CD, we measure the robustness of a model compared to the reference model. 
With rCD, we measure if a model gets less or more affected by the corruptions compared to their performance on clean images relative to the reference model.

\section{Experiments}

\textbf{Augmentations.} In our experiments, we use the released models from Detectron2 \cite{wu2019detectron2}, and MMDetection \cite{mmdetection} frameworks and codebase provided by the authors of YOLACT \cite{bolya2019yolact}. These instance segmentation models are often trained with relatively weaker augmentation techniques compared to object recognition tasks.
The models we use in our experiments are trained with only scale augmentation, in which images are resized by randomly sampled scales for accuracy improvements.
On the other hand, image recognition tasks use a broader range of data augmentation techniques, such as random crop and various versions of color jitter. Therefore, the pretrained models on the ImageNet dataset are trained with these additional augmentations.
In our analysis, we do not consider targeted augmentation techniques for any corruption and instead evaluate the models that are optimized to achieve the best validation score on the clean datasets.

\begin{figure}[]
  \centering
  \setlength\tabcolsep{0.25pt}
  \renewcommand{\arraystretch}{0.12}
  \begin{tabular}{ccc}
    & Clean & Speckle Noise  \\
    \rotatebox{90}{~~ BN pretrained} &  \interpfigi{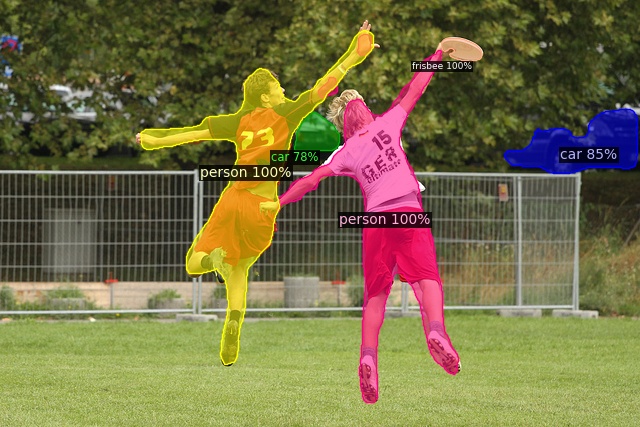} &
    \interpfigi{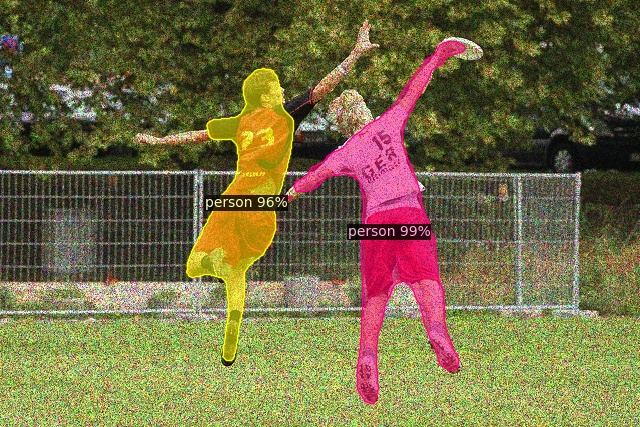}\\
    \rotatebox{90}{~SyncBN scratch}  &  \interpfigi{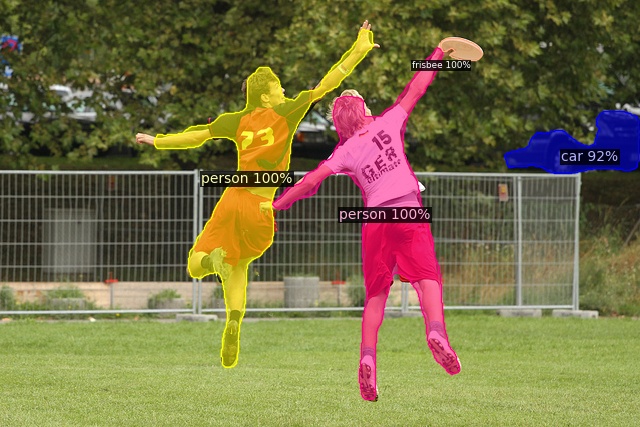}&
    \interpfigi{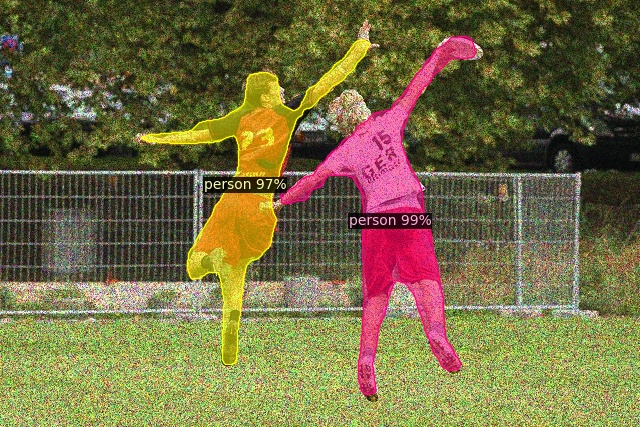}\\
    \rotatebox{90}{~~~GN scratch} & \interpfigi{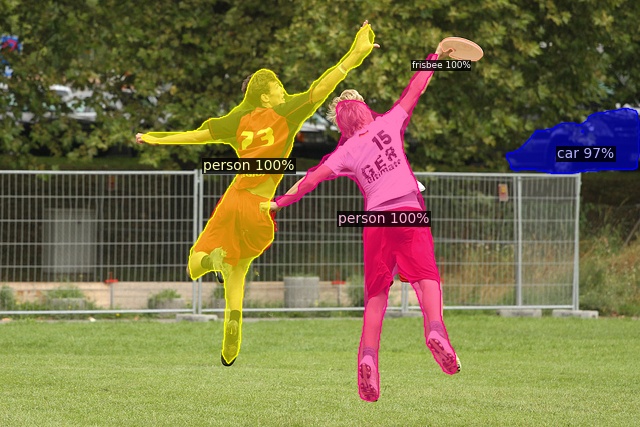}&
    \interpfigi{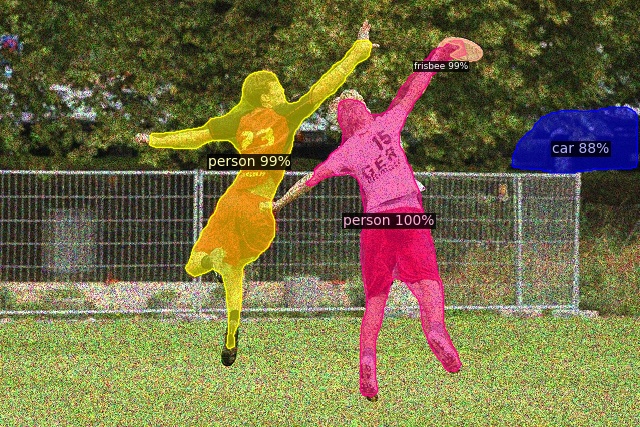}\\
     \end{tabular}
     \caption{Rows present the results of BN R50-FPN pretrained, SyncBN R50-FPN scratch, and GN R50-FPN scratch models, respectively.
     On the clean COCO validation image, models perform similarly.
     On the image corrupted with speckle noise, there are large performance gaps within models.
     Sync BN model from scratch outputs more accurate masks compared to BN pretrained model.
     GN from scratch model shows the highest robustness to the corruption and continues detecting the frisbee and the car with high confidence.}
     \label{fig:coco_norm}
\end{figure}

\begin{figure}[]
  \centering
  \setlength\tabcolsep{0.32pt}
  \renewcommand{\arraystretch}{0.15}
  \begin{tabular}{ccc}
    & Clean & Gaussian Noise  \\
    \rotatebox{90}{~~~~~~~~~~~~C4} &  
    \interpfigi{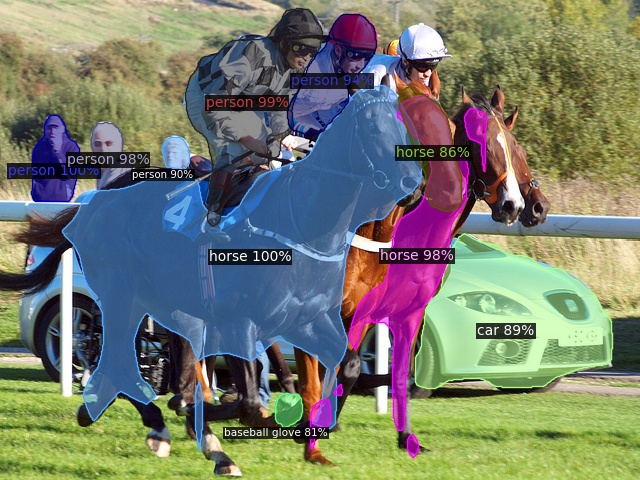} &
    \interpfigi{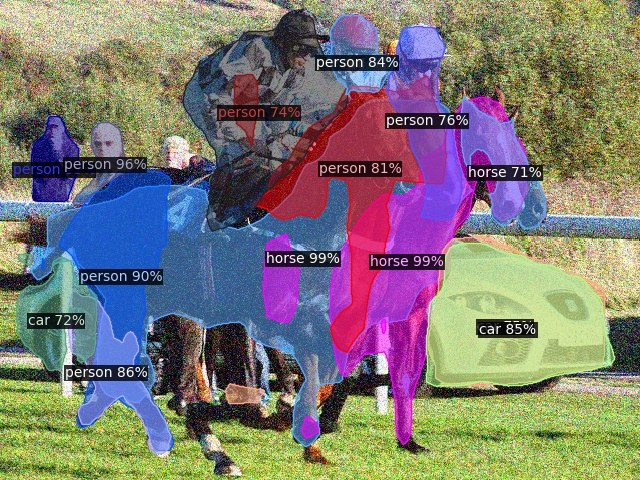}\\
    \rotatebox{90}{~~~~~~~~~DC5}  &  
    \interpfigi{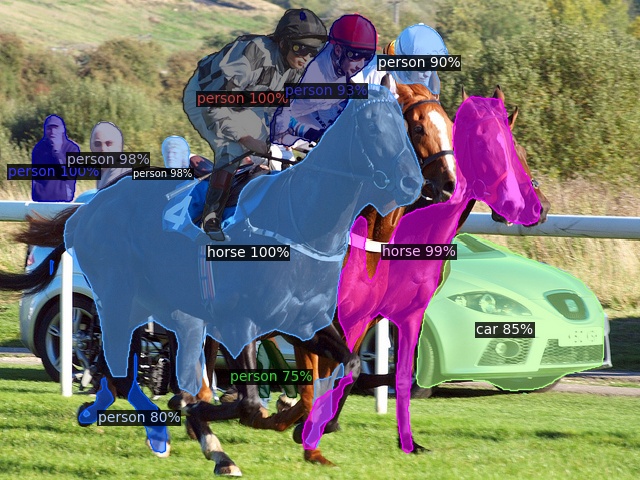}&
    \interpfigi{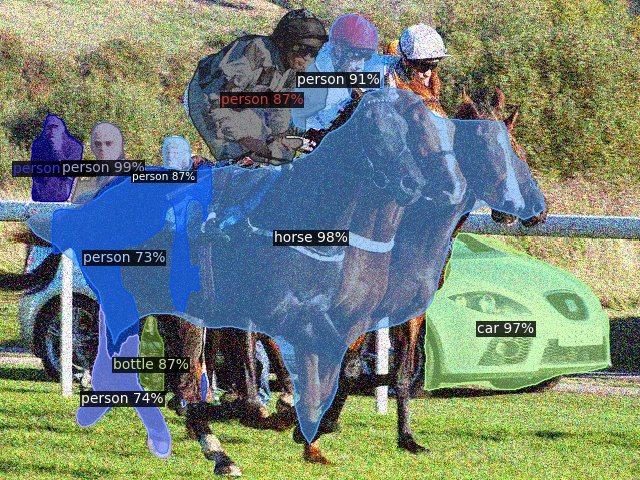}\\
    \rotatebox{90}{~~~~~FPN Cas.} & 
    \interpfigi{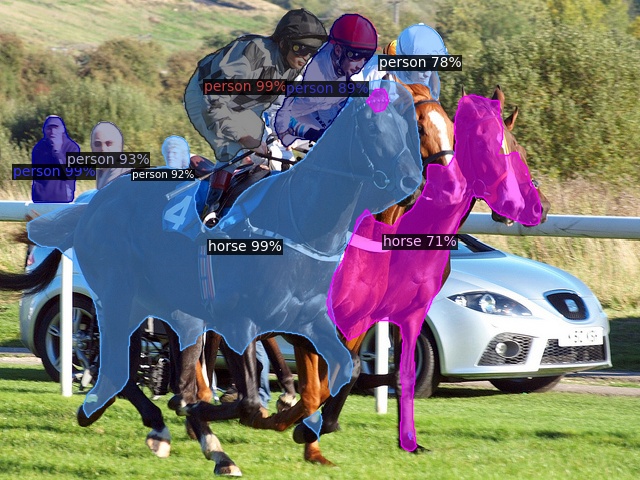}&
    \interpfigi{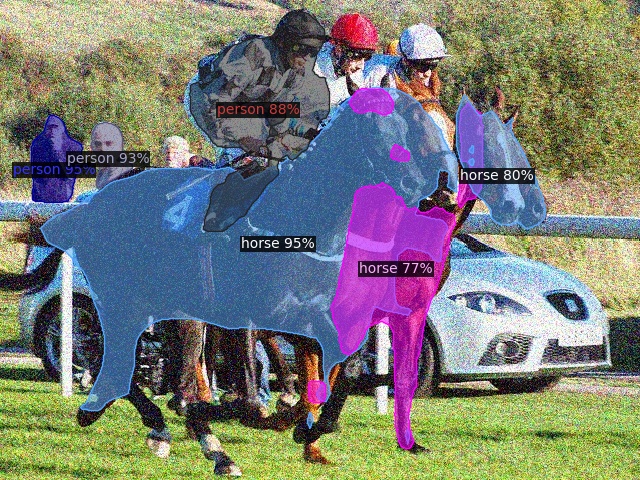}\\
    \rotatebox{90}{~~~FPN Deform.} & 
    \interpfigi{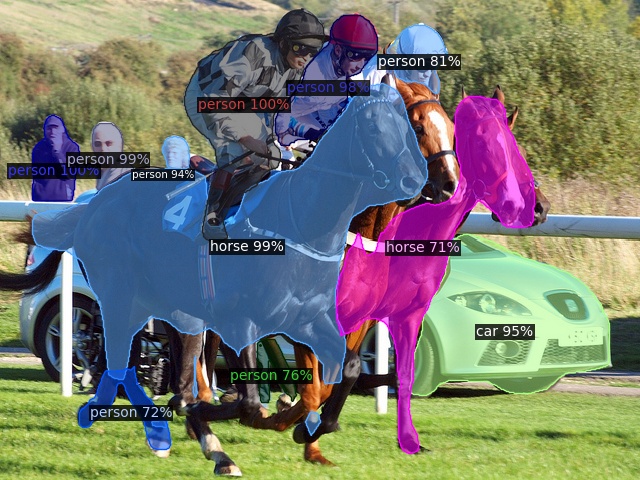} &
    \interpfigi{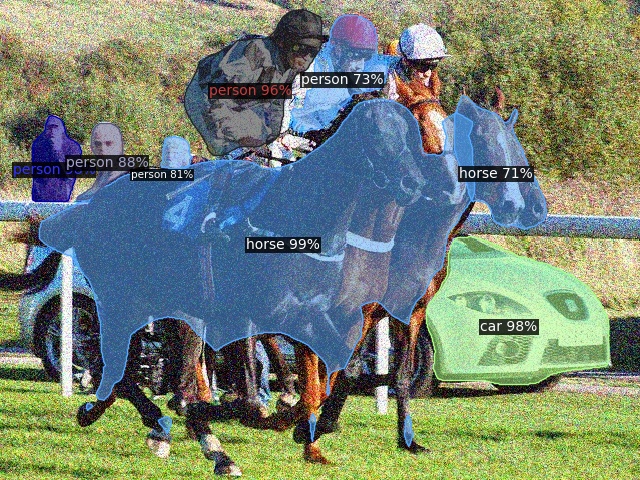}\\

     \end{tabular}
     \caption{Rows present the results of R50-C4, R50-DC5, R50-FPN Cascade models, and R50-FPN Deformable models, respectively.
     On the clean COCO validation image, models perform detections with high accuracy.
     On the other hand, on the Gaussian noise corrupted image, C4 architecture outputs noisy detections. DC5 based model achieves slightly better results but still outputs false detections such as the bottle.
     FPN deformable architectured model achieves the highest robustness in this example. 
     }
     \label{fig:coco_arch}
\end{figure}

\begin{figure}[t]
  \centering
  \setlength\tabcolsep{0.32pt}
  \renewcommand{\arraystretch}{0.15}
  \begin{tabular}{ccc}
    & Clean & Defocus Blur  \\
    \rotatebox{90}{~~~~~R101}  & 
    \interpfigi{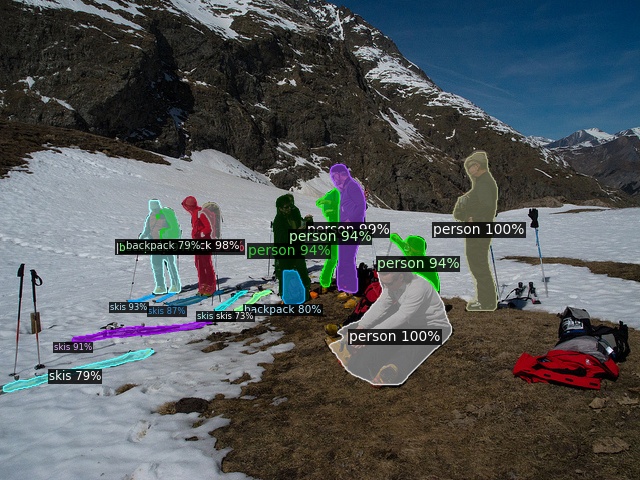}&
    \interpfigi{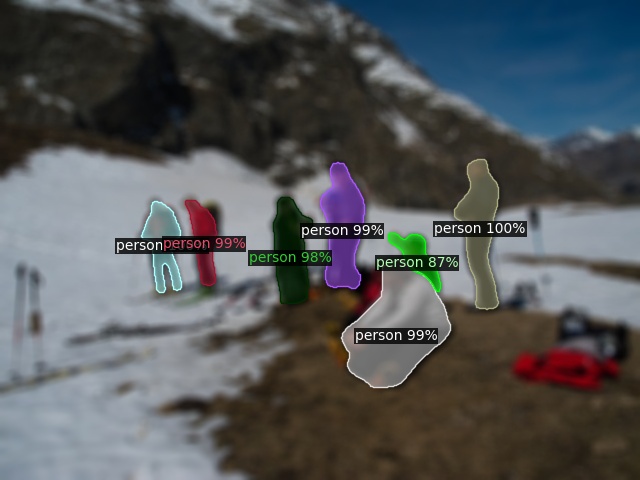}\\
    \rotatebox{90}{~~~~~X101} & 
    \interpfigi{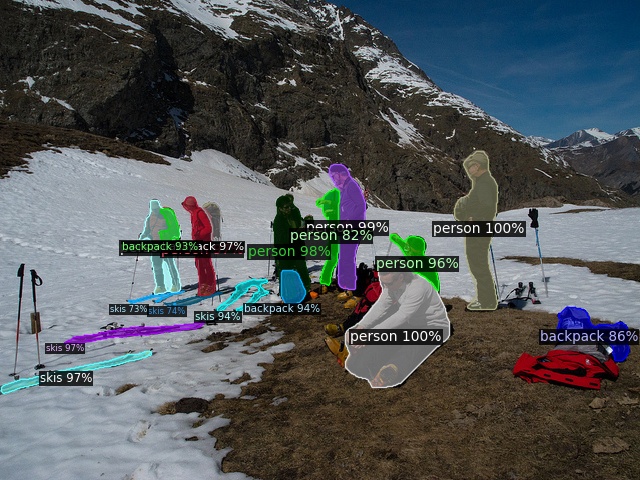}&
    \interpfigi{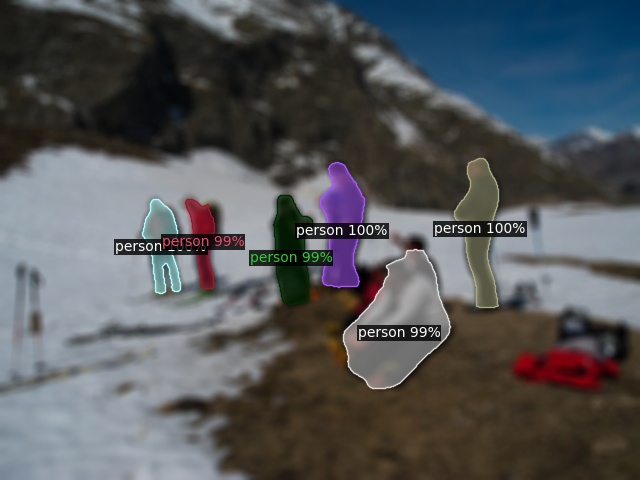}\\
    \rotatebox{90}{~~~~~S96} & 
    \interpfigi{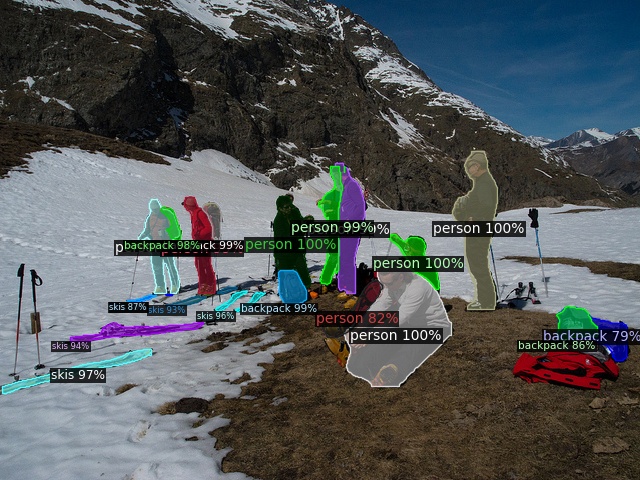}&
    \interpfigi{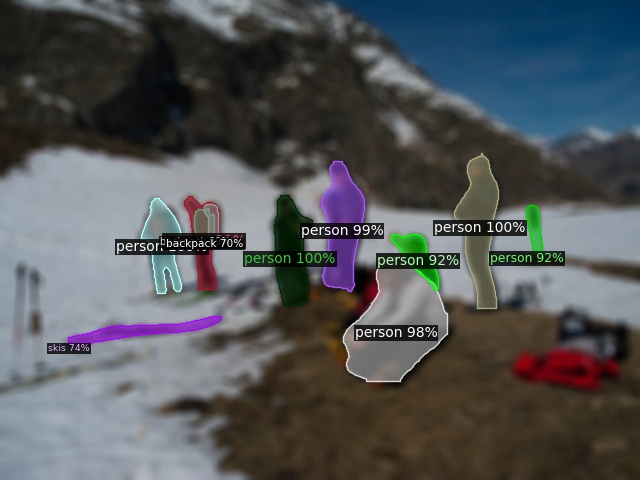}\\
     \end{tabular}
     \caption{Rows present results of R101-FPN, X101-FPN, and S96-FPN   models,  respectively.
     On clean images, models output comparable results.
     On defocus blur image, R101 and S96 can detect the highly occluded person (behind white), whereas X101 fails to detect.
     Furthermore, S96-FPN impressively detects the backpack and the ski on the challenging defocus blur image.}
     \label{fig:coco_backbone}
\end{figure}

\textbf{Training parameters.}  The same parameters are used following the author's released code. 
They are optimized to achieve the best validation scores by the authors.
For the architecture studies,  Mask-RCNN based models (C4, DC5, FPN, FPN-Deform, FPN-Cas), PointRend, and SOLO (SOLO, SOLOv2) models use the same setting.
Specifically, models are trained on $8$ GPUs with $16$ images per batch for $270000$ steps (approximately 36 epochs). 
The base learning rate of $0.2$ is used with a learning rate that is decreased by $10$ at $21000$ and $25000$ iterations. Weight decay is set to $0.0001$ and momentum to $0.9$.
QueryInst is also trained for 36 epochs. The initial learning rate is set to $2.5\times 10^{-5}$, divided by 10 at the 27th epoch and 33th epoch, respectively.
TensorMask is trained for $ 540000$ steps with an initial learning rate of $0.2$ is decreased by $10$ at $480000$ and $520000$ iterations. 
PolarMask is trained for $12$ epochs.
YOLACT models (YOLACT, YOLACT++) are trained on a single GPU with batch size 8 for $800000$ iterations.
The initial learning is set to $0.001$ and decreased by $10$ at iteartions $280000$, $600000$, $700000$, and $750000$. 
Weight decay is set to $0.0005$ and momentum to $0.9$.
For network backbones comparisons, ResNet and ResNeXt models are trained the same as the previously explained Mask-RCNN based models.
During inference, images are resized such that their scale (shorter edge) is 800 pixels.
SwinTransformer models are trained with AdamW optimizer with a learning rate of $0.0001$ betas of $(0.9, 0.999)$, and weight decay of $0.05$ for $36$ epochs.
For inference, images are again resized for their shorted edge to be 800 pixels.
MViTV2-T is also trained for $36$ epochs with a learning rate of $0.00016$.
SpineNet models are trained for $350$ epochs from scratch with SGD optimizer and  a learning rate of $0.14$, momentum of $0.9$, and weight decay of $0.00004$.
The SpineNet-49 model infers on $640\times640$ image resolution whereas the SpineNet-96 model on $1024$ and the SpineNet-143 model on $1280$.

\newcommand{\interpfigc}[1]{\includegraphics[width=7.6cm,height=3.2cm]{#1}}
\begin{figure}[]
  \centering
  \setlength\tabcolsep{0.32pt}
  \renewcommand{\arraystretch}{0.15}
  \begin{tabular}{cc}
    \rotatebox{90}{~~~~~~~~~~~~~~~C4} &  
    \interpfigc{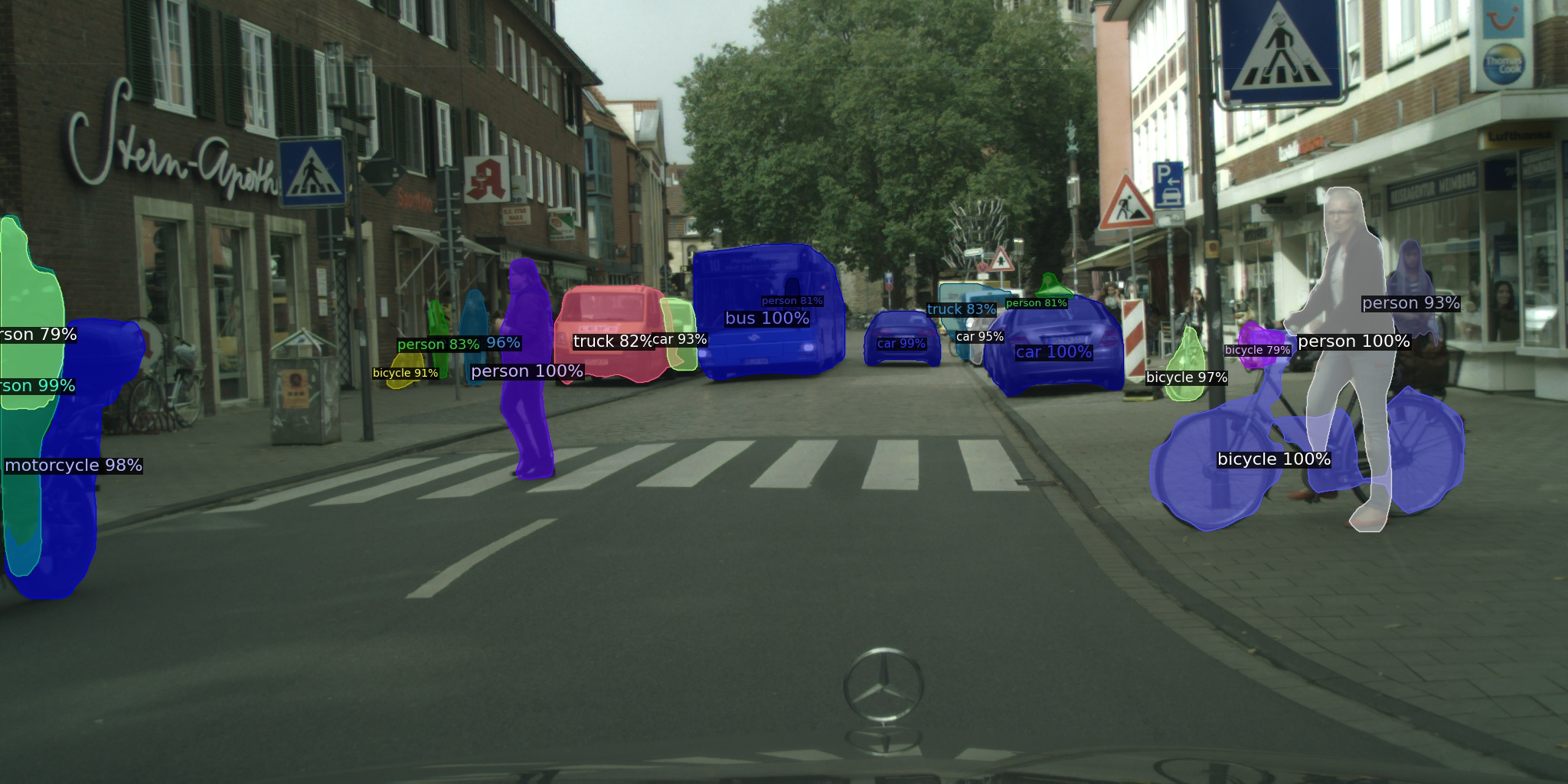} \\
    \rotatebox{90}{~~~~~~~~~~~~~DC5}  &  
    \interpfigc{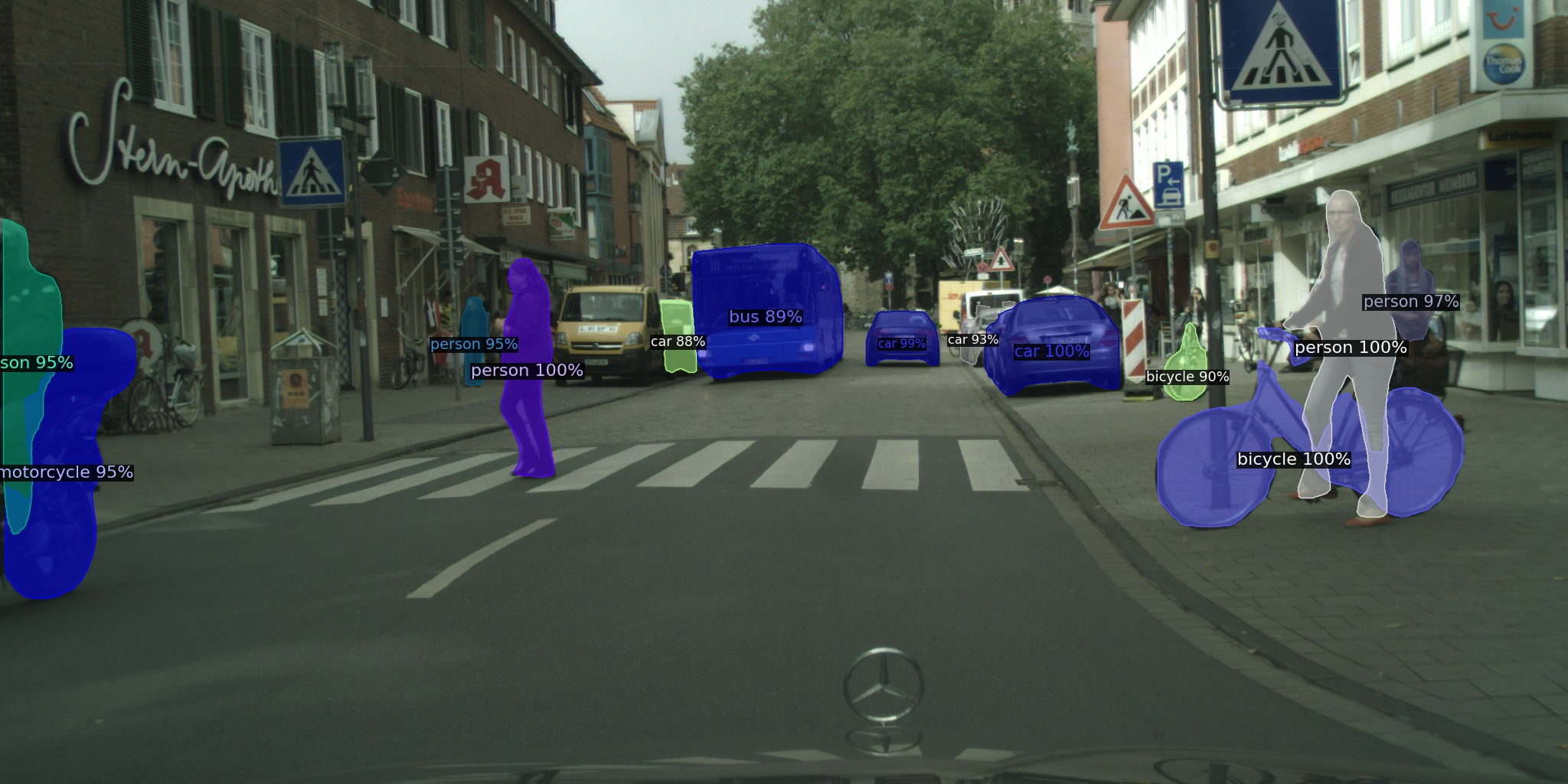} \\
    \rotatebox{90}{~~~~~~~~~~~FPN Cas.} & 
    \interpfigc{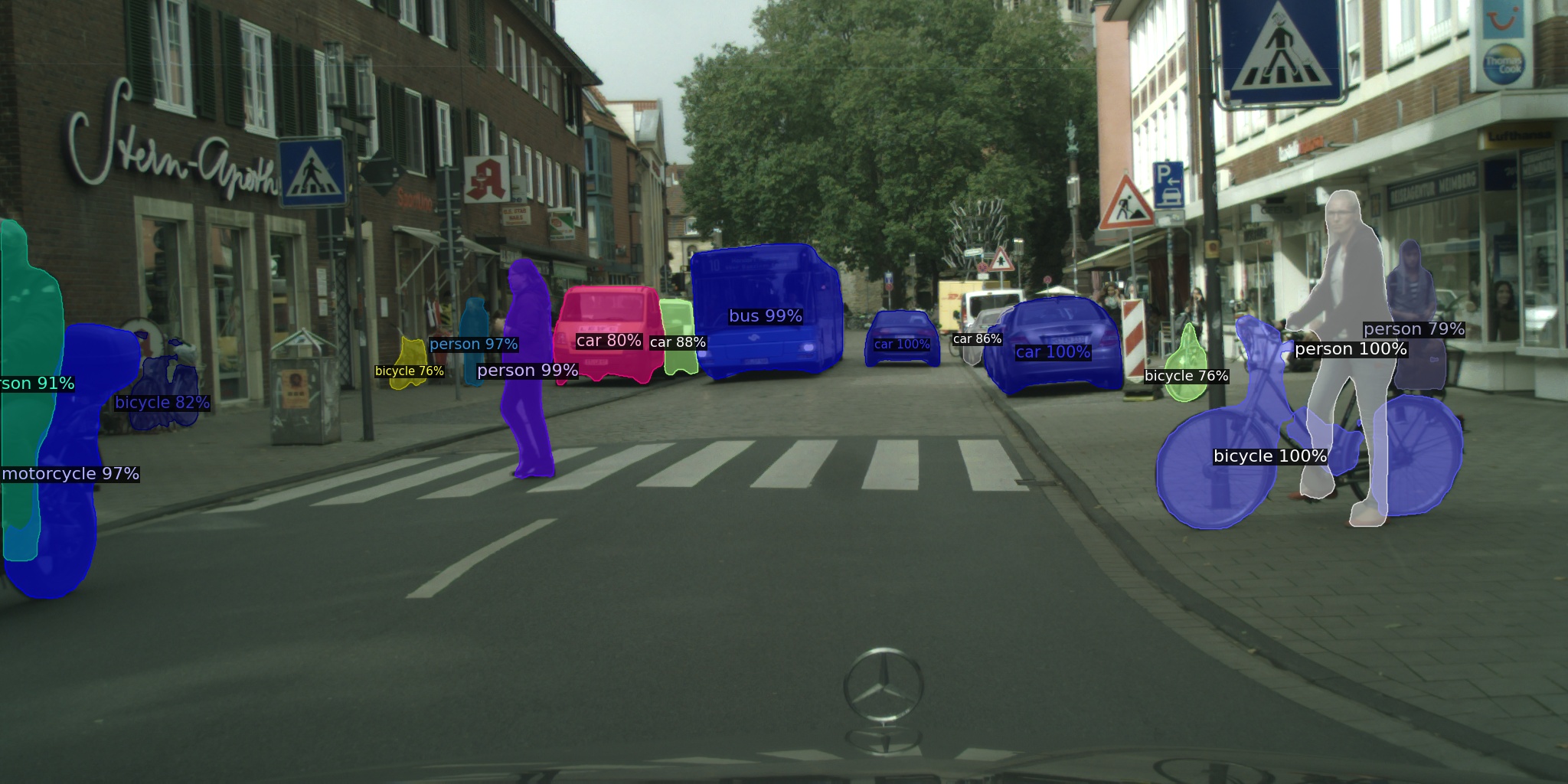} \\
     \end{tabular}
     \caption{Rows present results of R50-C4, R50-DC5, and R50-FPN Cascade models, respectively. Models are trained on the COCO dataset and inferred on Cityscapes images.
     C4 architecture outputs false detections of persons, and cars are misclassified as trucks.
     DC5 does not output false detections but misses true detections of cars and bicycles.
     Among these models, the FPN Cascade model detects all bicycles and the majority of the cars.}
     \label{fig:citys_noise}
\end{figure}

\begin{table*}[t]
\caption{Quantitative results on COCO validation dataset with images corrupted by various techniques. Instance segmentation scores are presented. The first column provides scores on the original (clean) validation dataset; the second column provides the results on the mixed corruption dataset (MC), the third presents the average of all scores on corrupted images (Avg.) with single corruptions. The fourth and fifth columns provide CD and rCD scores, respectively. rCD scores are based on reference models given in the first row of each table, and the rest of the columns show scores on the specified corrupted validation sets.}\label{tab:arch_all}

\begin{subtable}[h]{1.0\textwidth}
\centering
\renewcommand\tabcolsep{4pt}
\resizebox{\textwidth}{!}{
\begin{tabular}{l|c|c|c|c|c|ccc|cccc|cccc|cccc|c}
\toprule
& & &  & & & \multicolumn{3}{c}{Blur} & \multicolumn{4}{c}{Noise}  & \multicolumn{4}{c}{Digital}  & \multicolumn{4}{c}{Weather} & Geom. \\
\midrule
 & Clean & MC & Avg. & CD & rCD & Motion & Defoc. & Gaus. & Gaus. & Impulse & Shot & Speckle & Bright. & Contr. & Satur. & JPEG & Snow & Spatter & Fog & Frost & GTr. \\
\midrule
C4 & 34.3 & 7.5 & 19.7 & 100 & 100 & 12.8 & 14.7 & 15.8  & 14.6 & 12.5 & 15.1 & 18.3 & 30.2 & 21.7 & 32.3 & 16.1 & 15.9 & 23.5 & 25.9 & 17.2 & 29.4 \\
DC5 & 35.8 & 8.3 & 20.8 & 99 & 102 & 13.2 & 15.1 & 16.6  & 15.0 & 13.3 & 15.8 & 18.9 & 31.6 & 22.9 & 33.9 & 18.2 & 16.6 & 24.6 & 27.5 & 18.1 & 30.9 \\
FPN & 37.1 & 8.6 & 21.6 & 98 & 106 & 13.5 & 15.5 & 17.2  & 16.1 & 14.2 & 16.9 & 19.9 & 32.9 & 24.1 & 35.2 & 17.8 & 17.2 &  25.1 & 28.8 & 18.8 & 31.8 \\
FPN-Cas. & 38.4 & 9.1 & 22.4 & 97 & 109 & 13.7 & 16.0 & 18.0 & 17.1 & 14.9 & 18.0 & 21.0 & 33.9 & 25.2 & 36.4 & 18.7 & 18.3 & 25.9 & 29.9 & 19.9 & 32.7\\
FPN-Deform. & 38.5 & 10.6 & 23.5 & 95 & 102 & \textbf{16.1} & 18.0 & 19.5  & 18.0 & 16.0 & 18.7 & 22.0 & 34.8 & 25.9 & 36.6 & 19.9 & 19.1 & 26.4 & 30.7 & 21.1 & 33.2 \\
PointRend & 38.3 & 8.9 & 22.4 & 97 & 108 &  13.9 &  15.7 & 17.6  & 17.0 & 15.2 & 17.7 & 20.9 & 34.0 & 25.1 & 36.3 & 18.3 & 18.0 & 26.3 & 29.9 & 19.8 & 32.8 \\
QueryInst & \textbf{41.4} & \textbf{11.0} & \textbf{25.8} & \textbf{92} & 105 & 15.6 & \textbf{18.6} & \textbf{21.4} & \textbf{20.7} & \textbf{18.7} & \textbf{21.0} & \textbf{24.1} & \textbf{37.6} & \textbf{29.4} & \textbf{39.6} & \textbf{23.5} & \textbf{20.0} & \textbf{29.1} & \textbf{33.8} & \textbf{23.0} & \textbf{36.3}\\
\hline
TensorMask & 35.8 & 8.0 & 20.4 & 99 & 105 & 12.8 & 14.0 & 15.9 & 14.6 & 12.8 & 15.1 & 18.6 & 31.6 & 23.1 & 33.8 & 17.0 & 16.8 & 23.5 & 27.9 & 18.2 & 30.5\\
SOLO & 35.8 & 9.1 & 21.5 & 98 & 97 & 13.6 & 14.9 & 17.1 & 16.7 & 15.1 & 17.1 & 20.2 & 32.0 & 24.4 & 33.7 & 18.4 & 17.2 & 24.9 & 28.6 & 18.9 & 31.4 \\
SOLOv2 & \textbf{37.5} & 9.6 & 22.9 & 96 & 98 & 14.7 & 16.4 & 18.7 & 17.6 & 15.7 & 18.3 & 21.6 & \textbf{33.9} & 25.7 & \textbf{35.7} & 19.8 & 18.2 & \textbf{26.6} & 29.8 & 20.1 & \textbf{33.0} \\
PolarMask & 29.3 & 7.7 & 17.4 & 103 & 82 & 11.6 & 12.6 & 14.1 & 13.4 & 11.4 & 14.1 & 16.3 & 26.2 & 20.1 & 27.6 & 15.1 & 13.5 & 19.4 & 23.3 & 14.9 & 25.0 \\
YOLACT & 28.0 & 11.3 & 19.3 & 101 & \textbf{57} & 13.7 & 15.2 & 17.3 & 16.3 & 15.1 & 16.5 & 17.8 & 26.1 & 23.1 & 27.1 & 20.0 & 14.8 & 19.6 & 24.7 & 16.0 & 25.4 \\
YOLACT++ & 33.7 & \textbf{16.3} & \textbf{25.0} & \textbf{94} & \textbf{57} & \textbf{18.3} & \textbf{20.4} & \textbf{22.6} & \textbf{22.1} & \textbf{21.2} & \textbf{22.6} & \textbf{24.2} & 31.9 & \textbf{29.2} & 32.9 & \textbf{25.7} & \textbf{20.3} & 25.6 & \textbf{30.7} & \textbf{22.1} & 30.9 \\
\bottomrule
\end{tabular}}
\caption{Networks trained with different network architectures using R50 backbone. Multi-stage and single-stage instance segmentation models are separated by a horizontal line.}\label{tab:arch}
\end{subtable}

\begin{subtable}[h]{1.0\textwidth}
\centering
\renewcommand\tabcolsep{4pt}
\resizebox{\textwidth}{!}{
\begin{tabular}{l|c|c|c|c|c|ccc|cccc|cccc|cccc|c}
\toprule
& & & & & & \multicolumn{3}{c}{Blur} & \multicolumn{4}{c}{Noise}  & \multicolumn{4}{c}{Digital}  & \multicolumn{4}{c}{Weather} & \multicolumn{1}{c}{Geom.} \\
\midrule
 & Clean & MC & Avg. & CD & rCD & Motion & Defoc. & Gaus. & Gaus. & Impulse & Shot & Speckle & Bright. & Contr. & Satur. & JPEG & Snow & Spatter & Fog & Frost & GTr. \\
\midrule
R50 & 37.1 & 8.6 & 21.6 & 100 & 100 & 13.5 & 15.5 & 17.2  & 16.1 & 14.2 & 16.9 & 19.9 & 32.9 & 24.1 & 35.2 & 17.8 & 17.2 & 25.1 & 28.8 & 18.8 & 31.8 \\
R101 & 38.6 & 9.9 & 23.3 & 98 & 98 & 15.2 & 16.9 & 19.0 & 18.3 & 15.8 & 18.8 & 21.5 & 34.5 & 26.2 & 36.9 & 18.4 & 19.3 & 27.1 & 30.9 & 20.8 & 33.3 \\
X101 & 39.5 & 9.4 & 23.3 & 98 & 107 & 14.3 & 15.8 & 17.8 & 17.0 & 15.3 & 17.3 & 20.9 & 35.4 & 25.9 & 37.4 & 20.0 & 19.7 & 30.6 & 21.3 & \textbf{30.6} & 34.0 \\
S49 & 37.8 & 15.5 & 26.0 & 94 & 75 & 19.0 & 22.5 & 25.0 & 22.4 & 21.6 & 22.6 & 25.2 & 35.1 & 28.1 & 36.6 & 24.2 & 19.3 & 27.1 & 31.6 & 22.7 & 33.9 \\
S96 & 41.2 & 17.8 & 29.2 & 90 & \textbf{74} & 19.7 & 22.7 & 25.3 & 25.1 & 25.2 & 25.3 & 27.9 & 38.6 & 33.3 & 40.2 & 27.5 & 24.7 & 32.3 & 36.1 & 26.5 & 37.3 \\
S143 & 41.3 & 17.3 & 29.2 & 90 & 75 & 18.9 & 22.7 & 25.4 & 24.9 & 24.4 & 25.0 & 27.1 & 38.6 & 34.3 & 39.9 & 28.0 & 24.4 & 33.5 & 36.6 & 26.8 & 37.0 \\
SwinT & 41.7 & 15.1 & 28.5 & 91 & 84 & 20.4 & 22.3 & 24.8 & 22.9 & 24.6 & 23.3 & 25.9 & 38.4 & 33.0 & 40.1 & 24.7 & 23.9 & 32.9 & 36.3 & 25.6 & 36.9\\
SwinS & 43.2 & 17.2 & 31.0 & 88 & 76 & 21.9  & 23.9 & \textbf{26.9} & \textbf{26.3} & \textbf{27.5} & 26.5 & 29.1 & 40.2 & 35.7 & {42.0} & 28.9 & 26.7 & 35.2 & 38.5 & 27.7 & 38.8\\
MViTV2-T & \textbf{43.8} & \textbf{19.0} & \textbf{31.7} & \textbf{87} & 76 & \textbf{22.4} &  \textbf{24.3} &  \textbf{26.9} & 26.2 & 27.2 & \textbf{26.7} & \textbf{30.1} & \textbf{41.0} & \textbf{37.2} & \textbf{42.3} & \textbf{29.1} & \textbf{28.8} & \textbf{36.0} & \textbf{39.6} & 30.2 & \textbf{39.2} \\
\bottomrule
\end{tabular}}
\caption{Networks trained with different network backbones using FPN architecture.}\label{tab:backbone}
\end{subtable}
\begin{subtable}[h]{1.0\textwidth}
\centering
\renewcommand\tabcolsep{4pt}
\resizebox{\textwidth}{!}{
\begin{tabular}{l|c|c|c|c|c|ccc|cccc|cccc|cccc|c}
\toprule
& & & & &  & \multicolumn{3}{c}{Blur} & \multicolumn{4}{c}{Noise}  & \multicolumn{4}{c}{Digital}  & \multicolumn{4}{c}{Weather} & \multicolumn{1}{c}{Geom.} \\
\midrule
& Clean & MC &  Avg. & CD & rCD & Motion & Defoc. & Gaus. & Gaus. & Impulse & Shot & Speckle & Bright. & Contr. & Satur. & JPEG & Snow & Spatter & Fog & Frost & GTr. \\
\midrule
R50-FPN & 37.1 & 8.6 & 21.6 & 100 & 100 & 13.5 & 15.5 & 17.2 & 16.1 & 14.2 & 16.9 & 19.9 & 32.9 & 24.1 & 35.2 & 17.8 & 17.2 &  25.1 & 28.8 & 18.8 & 31.8 \\
GN R50-FPN & 38.6 & 11.5 & 24.8 & 96 & 87 & 16.2 & 15.4 & 18.1  & 19.7 & 18.6 & 20.2 & 23.1 & 35.0 & 34.4 & 36.9 & \textbf{22.6} & 20.5 & 27.7 & 32.8 & 21.6 & 33.6 \\
SyncBN R50-FPN & 37.8 & 8.7 & 22.1 & 99 & 100 & 13.3 & 15.3 & 17.1 & 15.6 & 13.9 & 16.1 & 19.7 & 33.5 & 27.1 & 36.0 & 19.3 & 17.9 & 25.7 & 30.7 & 19.5 & 32.3 \\
GN R50-FPN scr. & \textbf{39.6} & \textbf{14.0} & \textbf{26.3} & \textbf{94} & \textbf{82} & \textbf{18.0} & \textbf{17.5} & \textbf{20.3} & \textbf{20.7} & \textbf{20.8} & \textbf{21.1} & \textbf{23.7} & \textbf{36.5} & \textbf{36.2} & \textbf{38.1} & 19.9 & \textbf{24.0} & \textbf{30.7} & \textbf{35.0} & \textbf{24.7} & \textbf{34.4}  \\
SyncBN R50-FPN scr. & 39.3 & 9.6 & 22.8 & 98 & 106 & 13.7 & 15.7 & 17.8 & 16.7 & 16.0 & 17.3 & 21.0 & 34.9 & 26.6 & 37.2 & 14.8 & 19.5 & 27.6 & 31.3 & 21.3 & 33.4 \\
\bottomrule
\end{tabular}}
\caption{Networks trained with different normalization methods, pretrained versus from scratch trainings.}\label{tab:scratch}
\end{subtable}
\begin{subtable}[h]{1.0\textwidth}
\centering
\renewcommand\tabcolsep{4pt}
\resizebox{\textwidth}{!}{
\begin{tabular}{l|c|c|c|c|c|ccc|cccc|cccc|cccc|c}
\toprule
& & & & & & \multicolumn{3}{c}{Blur} & \multicolumn{4}{c}{Noise}  & \multicolumn{4}{c}{Digital}  & \multicolumn{4}{c}{Weather} & \multicolumn{1}{c}{Geom.} \\
\midrule
& Clean & MC & Avg. & CD & rCD & Motion & Defoc. & Gaus. & Gaus. & Impulse & Shot & Speckle & Bright. & Contr. & Satur. & JPEG & Snow & Spatter & Fog & Frost & GTr. \\
\midrule
Baseline - 108 ep. & 39.3 & 9.6 & 22.8 & 100 & 100 & 13.7 & 15.7 & 17.8 & 16.7 & 16.0 & 17.3 & 21.0 & 34.9 & 26.6 & 37.2 & 14.8 & 19.5 & 27.6 & 31.3 & 21.3 & 33.4 \\
Aug - 100 ep. & 40.3 & 12.6 & 25.1 & 97 & 94 & 16.3 & 18.8 & 21.4 & 20.3 & 18.8 & 20.6 & 23.5 & 36.1 & 28.7 & 38.2 & 17.4 & 21.4 & 29.5 & 32.6 & 23.3 & 34.2\\
Aug - 200 ep. & 41.7 & \textbf{13.0} & 26.2 & 96 & \textbf{93} & \textbf{17.3} & 20.0 & \textbf{22.8} & \textbf{20.7} & 19.0 & 21.0 & \textbf{24.2} & 37.8 & \textbf{30.6} & 39.8 & 17.7 & \textbf{22.2} & \textbf{31.1} & 34.2 & \textbf{24.6} & 36.0 \\
Aug - 400 ep. & \textbf{42.5} & 12.7 & \textbf{26.4} & \textbf{95} & 96 & 17.2 & \textbf{20.2} & \textbf{22.8} & 20.6 & \textbf{19.5} & \textbf{21.1} & 24.0 & \textbf{38.5} & 30.4 & \textbf{40.7} & \textbf{18.5} & 22.1 & \textbf{31.1} & \textbf{34.8} & 24.4 & \textbf{36.8} \\
\bottomrule
\end{tabular}}
\caption{Networks trained from scratch with and without strong augmentations with longer trainings with SyncBN 50-FPN scr. architecture set-up.}\label{tab:aug}
\end{subtable}

\end{table*}

\begin{table}[t]
\caption{ Quantitative results on COCO, Cityscapes validation and training sets, and BDD validation set are presented. 
rCD is calculated by taking the first row model as a reference, and the COCO validation score as the clean score.} \label{tab:cross_all}
\begin{subtable}[h]{0.48\textwidth}
\centering
\resizebox{0.9\textwidth}{!}{
\begin{tabular}{l|c|c|c|c|c|c}
\toprule
 & COCO & City. val  & City. train & BDD  & CD & rCD\\
\midrule
C4 & 34.3 &  26.0 & 25.3 &  19.4 & 100 & 100 \\
DC5 & 35.8  & 27.8 & 26.5 & 21.5 & 98 & \textbf{99} \\
FPN & 37.1  & 26.9  & 25.8 & 21.8 & 98 & 117\\
FPN-Cas. & 38.4  & 27.9  & \textbf{27.6}  & \textbf{24.8} &\textbf{96}& 113 \\
FPN-Deform. &{38.5}  & 27.2  & 26.1 & 20.9 & 98 & 131 \\
PointRend & 38.3  & 28.3 & 27.1 &  24.0 & 96 & 114 \\
QueryInst &  \textbf{41.4} & \textbf{29.5} & 27.0 &  23.8 & 96 & 140\\
\hline
TensorMask & 35.8 & \textbf{26.9} & \textbf{26.3} & \textbf{22.2} & \textbf{98} & \textbf{101} \\
SOLO & 35.8 &  20.1 & 20.0 &  17.2 & 106 & 163 \\
SOLOv2 & \textbf{37.5} & 23.5 &  23.3 &  18.8 & 102 & 151 \\
PolarMask & 29.3 & 17.1 & 17.2 & 15.0 & 109 & 126 \\
YOLACT & 28.0 &  10.4 & 10.4 &  9.9 & 118 & 176 \\
YOLACT++ & 33.7 &  14.9 & 14.2 & 12.8 & 113 & 194\\
\bottomrule
\end{tabular}
}

\caption{Networks trained with different network architectures with R50. Multi-stage and single-stage instance segmentation models are separated by a horizontal line. }\label{tab:arch_cross}
\end{subtable}

\begin{subtable}[h]{0.48\textwidth}
\centering
\resizebox{0.9\textwidth}{!}{
\begin{tabular}{l|c|c|c|c|c|c}
\toprule
 & COCO & City. val  & City. train &  BDD &  CD & rCD\\
\midrule
R50 & 37.1 & 26.9  & 25.8  & 21.8 & 100 & 100 \\
R101 & 38.6  & 28.5 & 27.7  & 22.6 & 98 & 100 \\
X101 & 39.5  & 29.6  & 29.1  & 23.0 & 97 & 99 \\
S49 & 37.8  & 29.8  & 28.8  & 24.9 & 96 & \textbf{81} \\
S96 & 41.2 & 32.5 &  31.4  & \textbf{25.5} & \textbf{93} & 91\\
S143 & 41.3  & 31.7  & 30.2  & 24.8 & 95 & 100\\
SwinT & 41.7  & 30.1  & 29.1 &  23.0 & 96 & 116\\
SwinS  & {43.2} & \textbf{33.0} &  \textbf{31.8} &  25.2 & \textbf{93} & 106\\
MViTV2-T & \textbf{43.8}  & 30.1 &  29.6 &  25.2 & 95 & 127  \\
\bottomrule
\end{tabular}
}
\caption{Networks trained with different network backbones with FPN.}
\label{tab:bacbone_cross}
\end{subtable}
\begin{subtable}[h]{0.48\textwidth}
\centering
\resizebox{0.9\textwidth}{!}{
\begin{tabular}{l|c|c|c|c|c|c}
\toprule
 & COCO & City. val &  City. train  &BDD  &  CD & rCD\\
\midrule
R50-FPN & 37.1 & 26.9 &  25.8 &  21.8 & 100 & 100  \\
GN R50-FPN & 38.6  & 27.8 &  26.7 &  22.4 & 99 & 106 \\
SyncBN R50-FPN & 37.8 & 28.0 &  27.4 &  24.5 & 98 & \textbf{92} \\
GN R50-FPN scr. & \textbf{39.6} &  \textbf{29.8} & 28.6 &  24.0 &\textbf{96} & 98\\
SyncBN R50-FPN scr. & 39.3 & 29.2  & \textbf{29.1}  & \textbf{26.0} & \textbf{96} & \textbf{92} \\
\bottomrule
\end{tabular}
}
\caption{Networks trained with different normalization methods, pre-trained versus from scratch trainings (scr.).}\label{tab:scratch_cross}
\end{subtable}

\begin{subtable}[h]{0.48\textwidth}
\centering
\resizebox{0.9\textwidth}{!}{
\begin{tabular}{l|c|c|c|c|c|c}
\toprule
 & COCO & City. val  & City. train &  BDD  &  CD & rCD\\
\midrule
Baseline - 108 epoch & 39.3  & 29.2  & 29.1 &  26.0 & 100 & \textbf{100} \\
Aug - 100 epoch & 40.3  & 30.7 &  29.9 &  25.3 & 99 & 103\\
Aug - 200 epoch & 41.7  & 31.6 &  30.8 &  27.7 & 97 & 104 \\
Aug - 400 epoch & \textbf{42.5} &   \textbf{32.7} &  \textbf{31.8} &   \textbf{28.5} & \textbf{96} & 102\\
\bottomrule
\end{tabular}
}
\caption{Networks trained from scratch with and without strong augmentations with longer trainings with SyncBN 50-FPN scr. set-up.}\label{tab:aug_cross}
\end{subtable}

\end{table}

For the models that are not trained with copy-paste augmentation, we evaluate a converged model trained for 108 epochs. 
For the models that are trained with copy-paste augmentations, we compare the converged model trained for 400 epochs and the intermediate models that the checkpoints are obtained at epochs 100 and 200.
Experiments that start from  pretrained models are trained on $8$ GPUs with $16$ images per batch for $270000$ steps which correspond to approximately $36$ epochs on the COCO dataset, as mentioned previously. 
Models trained from scratch are trained for $3\times$ more number of epochs.
They are trained with $64$ images per batch and $202500$ iterations with a base learning rate of $0.08$. The learning rate is decreased by $10\times$ at iteration steps of $187500$ and $197500$.

\section{Results}
In this section, we share our results.
First of all, we find that the mean AP of the methods under clean dataset and corrupted datasets show significant differences, as can be seen from Table \ref{tab:arch_all}.
Among the studied corruptions, the models are relatively robust on the digital type corruptions (contrast, brightness, and saturation) and gemonetric transformations, which preserve texture to a certain point.
On others, performance is almost halved compared to the accuracy obtained on clean data.
Additionally, models trained on the COCO dataset obtain lower scores on the Cityscapes and BDD  datasets compared to models that are trained on Cityscapes and BDD datasets.
Among the inspected models, the drop in accuracy is more for some of the models and is investigated in detail in this section.

\begin{table*}[t]
\caption{Quantitative results on COCO validation dataset with images corrupted by various techniques. Bounding box detection scores of networks trained with and without instance segmentation heads. 
The first column provides scores on the original (clean) validation dataset; the second column provides the results on the mixed corruption dataset (MC), the third presents the average of all scores on corrupted images (Avg.) with single corruptions. The fourth and fifth columns provide CD and rCD scores, respectively. rCD scores are based on reference models given in the first row, and the rest of the columns show scores on the specified corrupted validation sets.}
\centering
\renewcommand\tabcolsep{4pt}
\resizebox{\textwidth}{!}{
\begin{tabular}{l|c|c|c|c|c|ccc|cccc|cccc|cccc|c}
\toprule
& & & &  & &\multicolumn{3}{c}{Blur} & \multicolumn{4}{c}{Noise}  & \multicolumn{4}{c}{Digital}  & \multicolumn{4}{c}{Weather} & \multicolumn{1}{c}{Geom.} \\
\midrule
 R50-FPN & Clean & MC & Avg. & CD & rCD & Motion & Defocus & Gaus. & Gaus. & Impulse & Shot & Speckle & Brightness & Contrast & Saturate & JPEG & Snow & Spatter & Fog & Frost & GTr. \\
\midrule
without Mask & 40.2 & 9.1 & 23.2 & 100 & 100 & 15.0 & 17.1 & 18.6 & 16.8 & 14.0 & 17.4 & 21.2 & 35.8 & 26.0 & 37.9 & 19.2 & 18.6 & 27.2 & 31.0 & 20.8 & 34.7 \\
with Mask & \textbf{41.0} & \textbf{9.8} & \textbf{24.2} & \textbf{99} & \textbf{99} &  \textbf{15.9} & \textbf{17.9} & \textbf{19.4} & \textbf{18.1} & \textbf{15.9} & \textbf{18.9} & \textbf{22.5} & \textbf{36.5} & \textbf{26.9} & \textbf{38.8} & \textbf{20.1} & \textbf{19.4} & \textbf{27.9} & \textbf{31.9} & \textbf{21.1} & \textbf{35.6} \\
\bottomrule
\end{tabular}}
\label{tab:multi_task}
\end{table*}

\begin{table}[t]
\caption{Quantitative results on COCO, Cityscapes validation, and Cityscapes training sets are presented.
Results show bounding box detection scores of networks trained with and without instance segmentation heads.}
\centering
\resizebox{0.47\textwidth}{!}{
\begin{tabular}{l|c|c|c|c|c|c}
\toprule
R50-FPN  & COCO & City. val &  City. train & BDD & CD & rCD\\
\midrule
without Mask & 40.2 &  32.7  & \textbf{30.8}  & 24.5 & \textbf{100} & \textbf{100} \\
with Mask & \textbf{41.0}  & \textbf{32.9} & 30.7  &  \textbf{24.9} &\textbf{100}& 107 \\
\bottomrule
\end{tabular}
}
\label{tab:multi_task_cross}
\end{table}

\textbf{Network Architectures.} Results of different instance segmentation architectures with a fixed backbone of R50 are given in Table \ref{tab:arch} for various image corruptions.
Improvements on clean images translate similarly but with a diminishing return to the corrupted images, as can be seen from rCD scores, which stay relatively constant and slightly degraded compared to the reference model on Mask-RCNN variants. Single-stage detectors, SOLO, and YOLACT are more robust to blur, noise, and weather corruptions.
It can be seen from our results that deformable convolution both improves the clean AP and the average of the corrupted image scores when comparing FPN versus FPN-Deform and YOLACT versus YOLACT++ (the main change is the deformable convolutions over YOLACT.)
However, it does not affect the rCD score.
Note that even though MC scores are usually in agreement with averaged corruption results, MC is significantly better for YOLACT++, even though on average QueryInst achieves a better score.
That is because MC usually showcases the results of the worst-case scenario.
Even though QueryInst achieves better results on average, YOLACT++ achieves consistently better scores for both blur and noise corruptions.
Therefore, when different corruptions are added together, the method that achieves better scores on the majority usually achieves better MC score.

On Cityscapes and BDD images, as given in Table \ref{tab:arch_cross}, the multi-stage models achieve significantly better scores than the single-stage instance segmentation models.
One important difference between  COCO images and these two datasets is that cross-dataset images are in much higher resolution, especially Cityscapes images have the size of $1024\times2048$. Network architectures are extensively tuned for different image resolutions for better accuracies.
However, in our experiments, we use the same evaluation settings proposed for COCO and do not change any parameters for cross-dataset experiments except the input image resolution which we use the original input resolutions without scaling. 
With this setup, single-stage architectures achieve worse results than the multi-stage instance segmentation models except for TensorMask, which achieves similar scores with its two-stage counterparts.
SOLO and SOLOv2 divide images into uniform grids, and if the center of an object falls into a grid cell, that grid cell is responsible for predicting the semantic category and masks of instances.
Therefore, grids should be small enough to contain only one instance and Cityscapes and BDD images are very crowded. 
Even though we do not target tuning any parameters for different datasets, we tried with a few options of grid size for SOLO and SOLOv2 but did not achieve better results. 
It is possible to adapt the inference parameters to achieve better scores on cross-datasets.
However, we are interested in using the model off-the-shelf on new image collections.
For that, SOLO and SOLOv2 do not seem to be good options. 
Similar issues happen with YOLACT versions. They do not work well on higher image resolutions.

Based on our evaluations, we conclude  Cascaded model  generalizes better to Cityscapes and BDD images.
The results get better from the single-stage to two-stage instance segmentation model, and the best results are obtained moving from two-stage to multi-stage (cascaded) architecture.
Visual results are presented for different architectures in Fig. \ref{fig:citys_noise}. 

\textbf{Network Backbones.} We measure the robustness of various backbone models in the FPN setting, as shown in Table \ref{tab:backbone}. We find that the network backbones affect robustness significantly.
On clean data, ResNet101 performs better than ResNet50, and ResNeXt101 achieves the best score among these three.
On the other hand, this ranking is not preserved for the corrupted validation datasets. On blur and noise corruptions, ResNet101 shows more robustness compared to ResNeXt101.
In contrast, the ResNeXt101 backbone results in better accuracies on digital and weather-type corruptions.
SpineNet, Swin, and MViTv2-T architectures  achieve significantly higher robustness to image corruptions and achieve very robust rCD scores as well as impressive clean AP scores.
These architectures have powerful representation capacities. 
Swin and  MViTv2-T architectures have global receptive fields due to their attention mechanism, whereas SpineNet architecture, due to its scale permuted architecture, has a very large receptive field.
We believe their global receptive field contributes to their robustness to image corruptions.
Visual results of this comparison are shown in Fig. \ref{fig:coco_backbone}.
On cross-dataset evaluations, as given in Table \ref{tab:bacbone_cross}, model performances on COCO, Cityscapes, and BDD show  positive correlations.
Original SpineNet inference set-ups process images on different resolutions as mentioned in Section \ref{sec:net_bone}. SpineNet-49 infers on $640\times640$ images, whereas the image size gradually increases for SpineNet-96 and SpineNet-143. If we use that original set-up when inferring on Cityscapes training images, methods achieve $18.1$, $23.8$, and $25.3$ scores, respectively. 
However, when we do not scale Cityscapes images, and run the inference on the original resolution, the scores increase to $21.8$, $31.4$, and $30.2$, respectively. S96 achieves slightly better results than S143 when it does not have the shortcoming of processing images in a lower resolution than S143.

We observe that, with the exception of S143, the larger models come with better Cityscapes and BDD scores.
For example, going from SwinT to SwinS, the number of parameters increases from 48M to 96M.
While SwinT achieves only $1.5$ AP better on the COCO validation set compared to SwinS, it achieves a $2.7$ AP better result on the Cityscapes training dataset, which has a much higher resolution.
MViTV2-T, on the other hand, has 44M parameters similar to SwinT and has improved architecture when measured on the COCO validation set.
However, its performance on the Cityscapes dataset is only slightly better than SwinT and worse than SwinS, which has a larger capacity.

\newcommand{\interpfigbd}[1]{\includegraphics[width=4.2cm]{#1}}
\begin{figure}[]
  \centering
  \setlength\tabcolsep{0.32pt}
  \renewcommand{\arraystretch}{0.15}
  \begin{tabular}{c|c}
    \interpfigbd{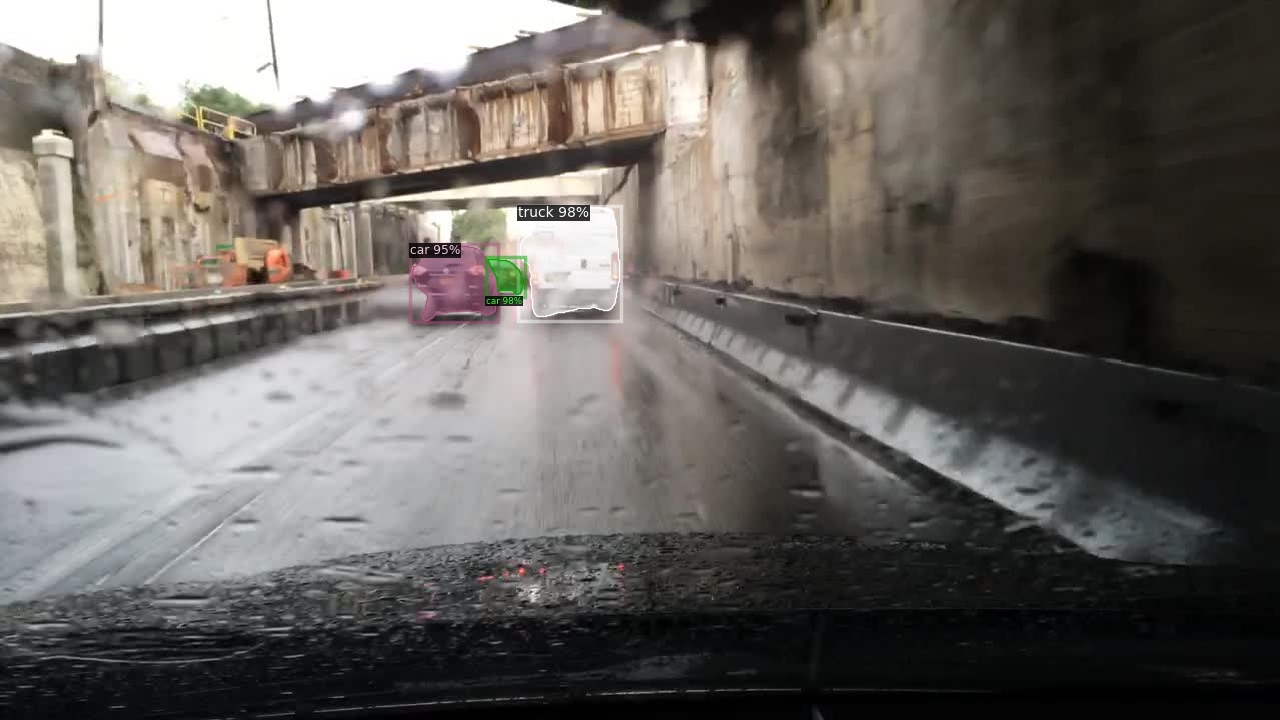} &
    \interpfigbd{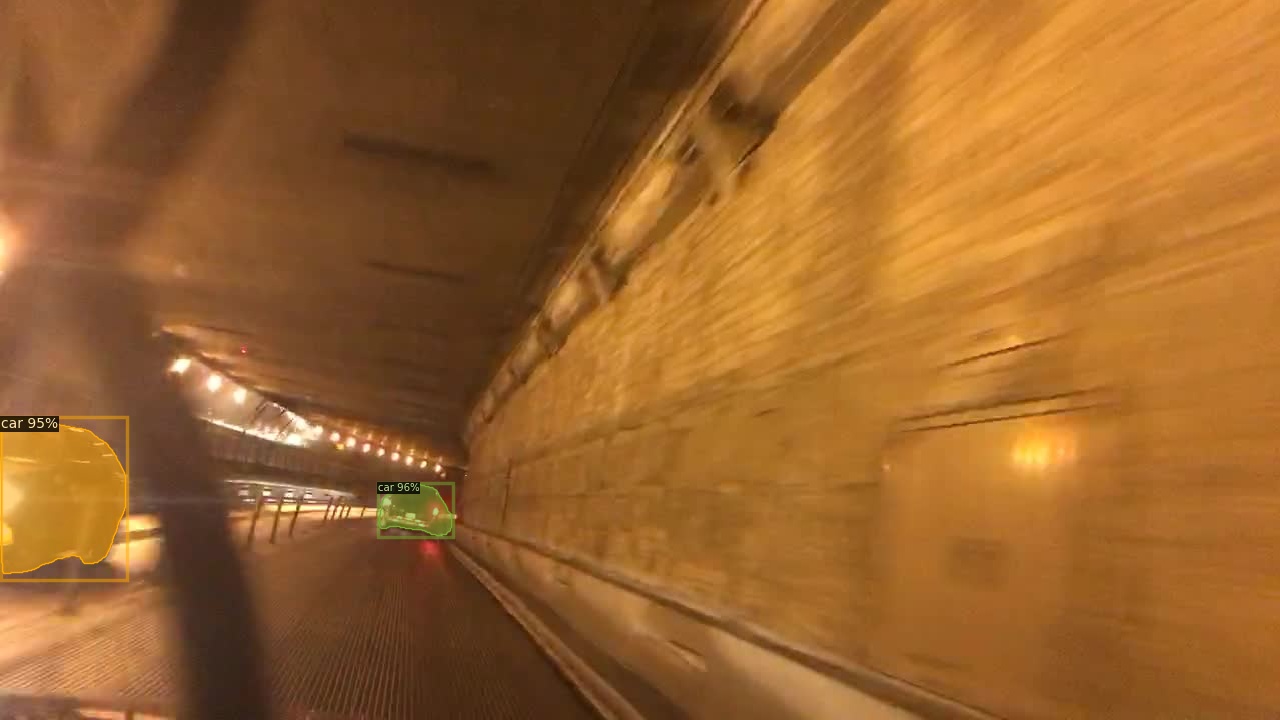} \\
    \interpfigbd{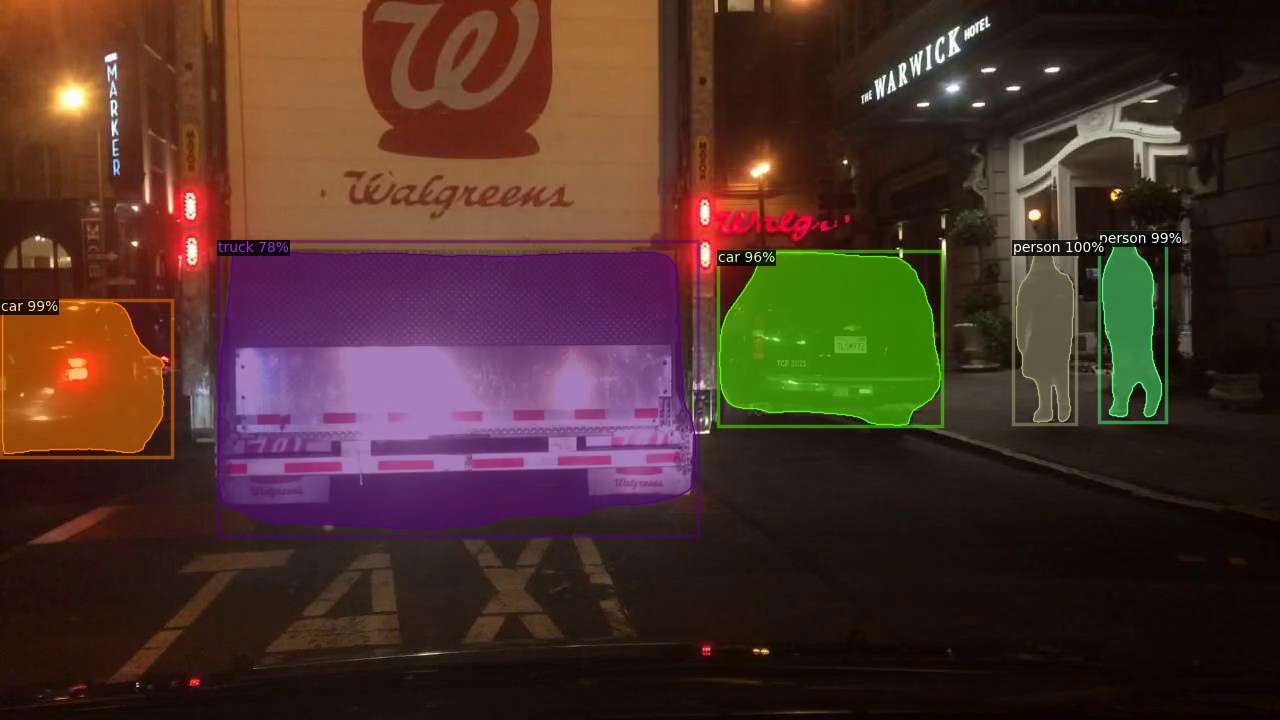} &
    \interpfigbd{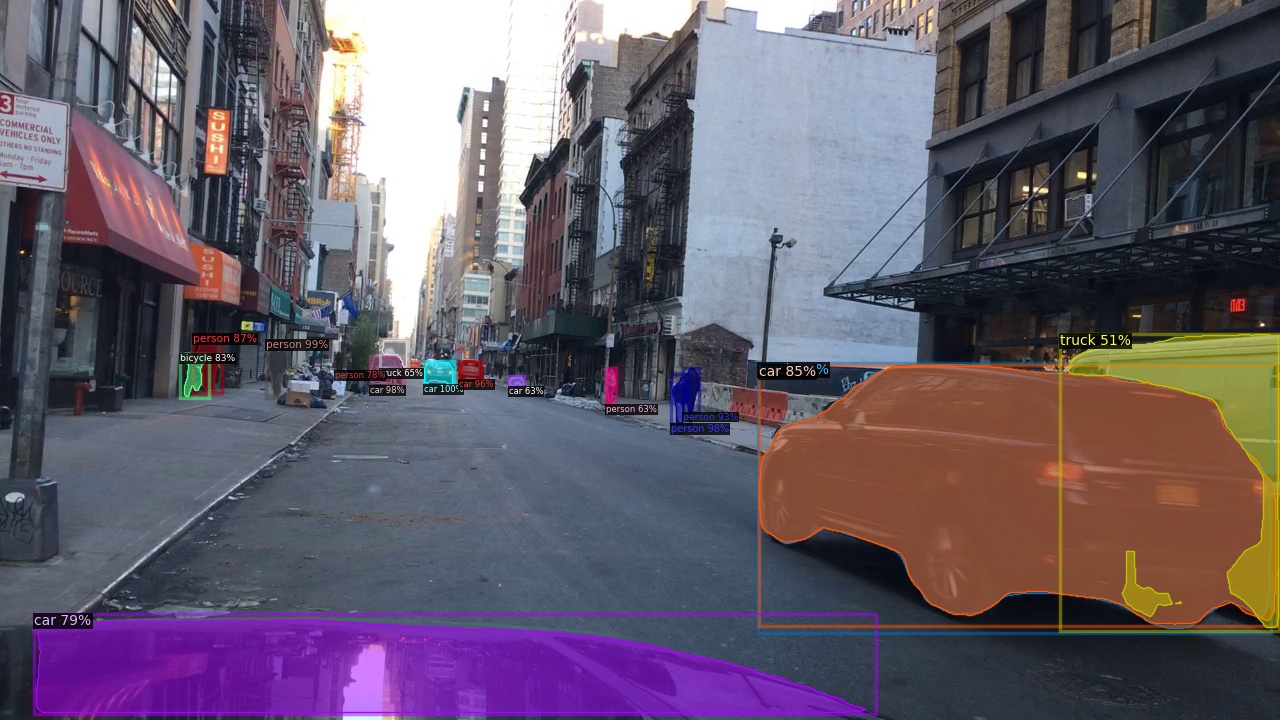}
     \end{tabular}
     \caption{Examples from BDD dataset together with the predictions from Aug. 400 network. BDD dataset includes diverse images as shown here from day time to night time to different weather conditions and even photos in tunnels.}
     \label{fig:bdd_noise}
\end{figure}

\textbf{ImageNet pretrained versus from scratch.} Recently, it has been shown that ImageNet pretraining does not improve accuracy on the COCO dataset, and even better accuracies can be obtained with training from scratch at the expense of longer training iterations \cite{he2019rethinking}.
Another interesting question is if ImageNet pretraining improves the robustness across different image corruptions. During pretraining, the networks learn from an additional million of images, which hypothetically can improve the robustness of learned features. Results of this comparison are shown in Table \ref{tab:scratch}.
Surprisingly, the models trained from scratch achieve higher accuracy on clean data and show significantly better robustness across different image corruptions except for JPEG corruption.
ImageNet pretrained models achieve significantly higher accuracies than their trained from scratch counterparts in JPEG corruption (22.65 vs. 19.89 for group normalization and 19.33 vs. 14.83 for synchronized batch normalization models).
On the other hand, among all other corruption types, models trained from scratch achieve higher robustness.
The same also holds on the Cityscapes and BDD evaluation, where we observe that models trained from scratch achieve higher accuracy results as given in Table \ref{tab:scratch_cross} on all sets.

\textbf{Normalization Layers.} Our results presented in Table \ref{tab:scratch} show that across normalization layers that are employed in instance segmentation models, group normalization provides increased robustness to networks under image corruptions.
This increase in robustness holds for both models trained from ImageNet pretrained networks and randomly initialized networks.
Interestingly, the improvements from group normalization compared to synchronized batch normalization are pretty significant for corrupted images than the clean images.
GN R50-FPN and SynchBN R50-FPN from scratch models perform closely on the clean datasets, $39.62$ and $39.26$, respectively.
On the other hand, looking at the corrupted image results on average, accuracy is $26.3$ and $22.8$, respectively, showing a large margin of improvements as can be measured by rCD scores.
Example output results are shown in Fig. \ref{fig:coco_norm}.

Results on Cityscapes and BDD images are quite interesting. Synchronized batch normalization models achieve significantly better results than group normalization models, as shown in Table \ref{tab:scratch_cross} showing an opposite behavior than the performance on corrupted images.
Even though GN-based models achieve a higher score in the COCO validation set, SyncBN-based models achieve higher scores on Cityscapes training datasets and BDD dataset.
Note that, there is a disagreement in the ranking of models between Cityscapes validation and training datasets, and we find the training dataset results to be more reliable given the larger number of samples it contains (2975 training images versus 500 validation images). 

This is a very interesting result. Adaptive normalization layers and domain-specific batch normalization layers are extensively studied for domain adaptation problems \cite{fan2021adversarially,chang2019domain, wang2019transferable}.
In this work, we are not interested in adapting one network to a target domain but interested in finding the most reliable network without additional tuning.
Both corruptions and cross-datasets change the statistics of the datasets. 
Corruptions change pixel-wise image statistics and cross-datasets change the content-based object statistics.
While corruptions do not change the content, on cross-datasets, objects appear in different densities and sizes. 
GN is better at making the network more robust to corruptions added onto images by removing their effects through normalization.
On the other hand, SyncBN is better at aligning the statistics of cross-datasets.

\textbf{Copy-Paste Augmentation.} It has been shown that copy-paste augmentation combined with longer training significantly improves performance on clean validation sets \cite{ghiasi2021simple}.
In Table \ref{tab:aug}, we compare a baseline that is not trained with copy-paste augmentation for 108 epochs and networks trained with copy-paste augmentations for 100, 200, and 400 epochs.
Note that the baseline trained for 108 epochs and the network trained with copy-paste augmentations for 400 epochs are the ones that converged (the validation set performance no longer improves).
Our results show that copy-paste augmentation significantly improves the robustness, especially early in training (100 epoch results).
Results on the cross-dataset experiments are mostly in agreement with the corruption scores, as shown in Table \ref{tab:aug_cross}.
Except for the slight decline in BDD scores from baseline to Aug 100 model, all scores improve consistently as the COCO score increases.
We provide visual predictions for Aug 400 model on BDD images in Fig. \ref{fig:bdd_noise}.

\textbf{Effect of Joint-task Training.} The results of this experiment are given in Table \ref{tab:multi_task} and \ref{tab:multi_task_cross}.
Both rows in tables share the same setup except that the model reported in the first rows is trained only to detect bounding boxes. In contrast, the second-row model learns to predict instance segmentation in addition to detecting bounding boxes.
They both use the ResNet50-FPN architecture. We compare the bounding box detection APs of these two models differing in terms of their prediction heads.
We find that the joint-task training improves the scores on the clean images and corrupted ones even more so than the improvement on clean images.
While the AP improvement coming from joint-task training is $0.77$ on clean images, it is $1.00$ on corrupted images on average and results in an rCD score of $99$.
Especially on noise corruptions, the improvements are pretty significant.
The results are similar on cross-dataset experiments as shown in Table \ref{tab:multi_task_cross}.
Even though without mask slightly outputs better than with a mask on the Cityscapes training dataset, on average, we observe better generalization of the method with joint-task training.

\newcommand{\interpfigt}[1]{\includegraphics[trim=0 0 0 0, clip, width=4.15cm]{#1}}
\begin{figure*}[t]
  \centering
  \begin{tabular}{cccc}
   \interpfigt{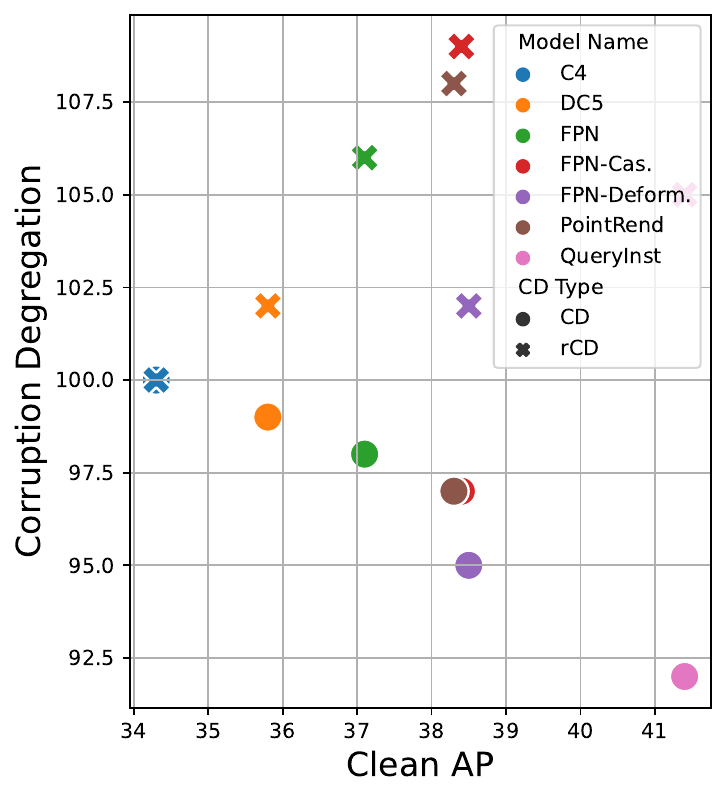}&
   \interpfigt{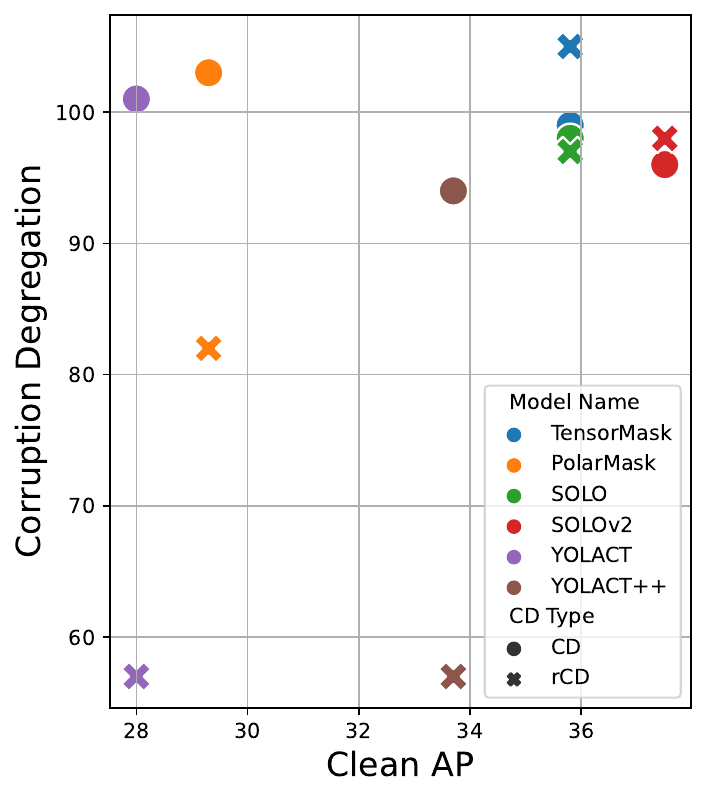}&
   \interpfigt{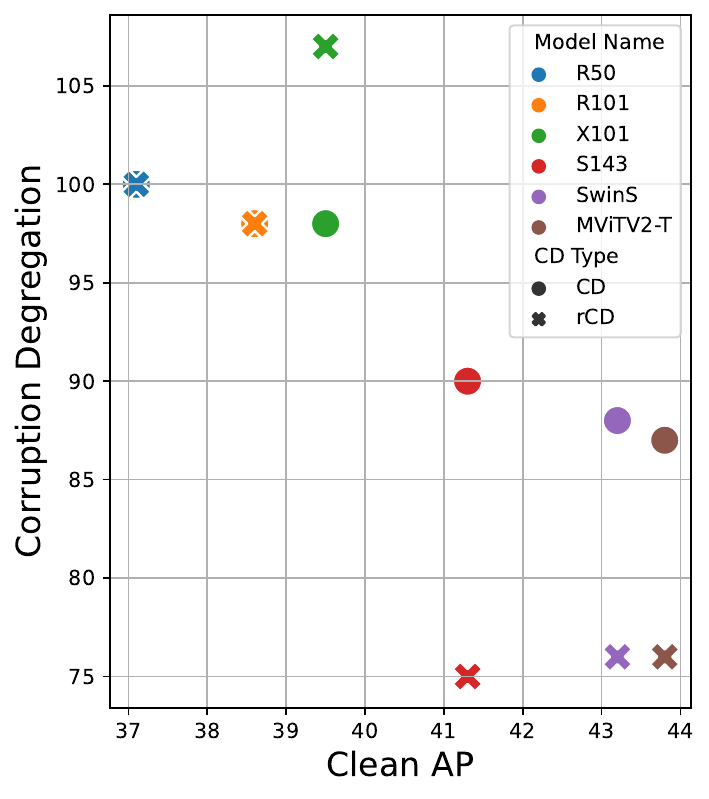}&
   \interpfigt{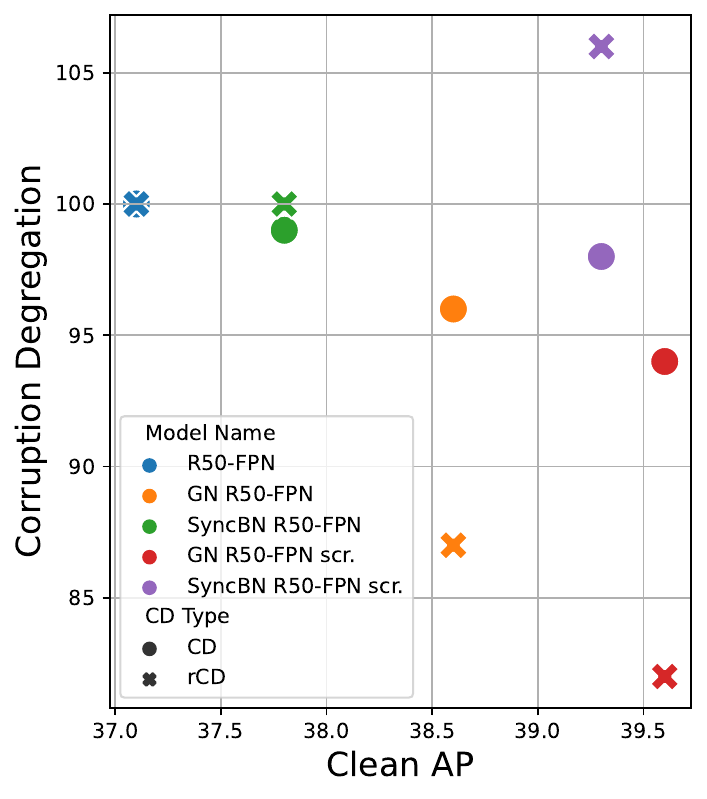}\\
    (a) & (b) &  (c) & (d)\\
     \end{tabular}
     \caption{CD and rCD for networks trained with  different architectures - multi-stage detectors (a),   - single-stage detectors (b), different backbones (c), different normalization layers and initializations (d)
     rCD and CD values below $100\%$ represent higher
     robustness than the reference model.
     Generally, when measured with CD, robustness improves (decrease of CD) with model performance on clean data.
     rCD shows a different behavior as usually improvements on the clean dataset are larger than the corrupted datasets hence rCD is higher than  $100\%$  in many cases.  }
     \label{fig:corruption_rcd}
\end{figure*}

\section{Discussion}
Researchers switch their attention to the robustness of deep models as they now achieve high-performance results on clean/in-domain validation sets.
Previously, the object recognition \cite{hendrycks2019benchmarking}, and semantic segmentation \cite{kamann2020benchmarking} tasks are rigorously evaluated for image corruptions. We are the first to conduct a similar study on the instance segmentation task to the best of our knowledge.
In addition to evaluating models under image corruptions, we also evaluate on images collected with a different set-up and from different environments than the training images; in other words, the images that show the domain gap with the training images.
We believe this is a very important addition to our study, which was not studied before among the previous robustness  papers \cite{hendrycks2019benchmarking, kamann2020benchmarking}.
One may argue that cross-dataset evaluations measure how reliable models are in a more realistic setting.
For models to be useful in real-world, they should be able to generalize well to different image resolutions and different sizes of objects and density of them in a scene. 
Our evaluation is also not limited to network architectures and backbones but extended to models with different normalizations, augmentations, initializations, and also investigates the impact of joint-task training. 

Through this study, we gain several insights.
In Fig.  \ref{fig:corruption_rcd}a, we compare different multi-stage network architectures and see that networks that achieve better clean AP scores also achieve better CD scores but with incremental returns, as can be measured by rCD.
On the other hand, on single-stage networks, as shown in Fig \ref{fig:corruption_rcd}b,  YOLACT variants achieve significantly more robust results.
For these novel architectures except YOLACT, it seems that architectures have slowly and consistently improved their representations over time. However, their corruption robustness improvements (CD) are explained mainly by better-learned representations on clean images as their rCD scores are larger than $100\%$.
On the other hand, YOLACT and YOLACT++ show a very different behavior. While they have lower clean AP scores, they achieve significantly better scores on corrupted images and an impressive rCD score, as can be seen in Fig. \ref{fig:corruption_rcd}b.
This behavior may be coming from its single-stage architecture.
However, on Cityscapes images and BDD images, single-stage detectors do not perform well.
They do not generalize well to larger image resolutions limiting their application to real-world use cases.

In our backbone studies, Fig.  \ref{fig:corruption_rcd}c, we find that deeper models and models with large receptive fields generally show higher robustness and generalization. The most significant results are from SpineNet and Swin backbones, which achieve below $80\%$ rCD scores on corrupted images.
Especially S49 achieves the best rCD score both on corrupted and cross-dataset images, $75\%$ and $81\%$, respectively.
The underlined difference regarding SpineNet architectures is an important finding, as previous benchmarking studies state that rCD scores are relatively constant among different network backbones \cite{hendrycks2019benchmarking, kamann2020benchmarking}.
Those studies did not include SpineNet architecture \cite{du2020spinenet}, which seems to have an improved architectural design for robustness.
Transformer-based architectures are similar, they also show improved architectural design for robustness and generalization.

We find that normalization layers affect robustness significantly.
As shown in Fig. \ref{fig:corruption_rcd}d, the better clean AP score of normalization layers correlates with a better CD.
Seeing such a positive correlation between clean AP and CD is expected, and the same observations are made for semantic segmentation tasks \cite{kamann2020benchmarking}.
The more interesting is the rCD score, which is reported to stay relatively constant in previous studies \cite{hendrycks2019benchmarking, kamann2020benchmarking}.
In Fig \ref{fig:corruption_rcd}d, we see that Group Normalization (GN) achieves a very robust rCD score.
The model shows larger improvements in robustness than it does for the clean validation set.
On the other hand, in cross-dataset experiments, we see that Synchronized Batch Normalization achieves better results.
This opens a new research direction for normalization layers.
When new normalization methods are proposed, it is best to evaluate them on clean validation sets as well as on corrupted sets and different domain images.
The trend so far is to either test a proposed normalization layer on clean validation datasets \cite{ulyanov2016instance, ioffe2015batch} or in domain adaptation setting on target datasets \cite{fan2021adversarially, chang2019domain} but not altogether, which we hope our work will inspire to change.
We also compare models trained from pretrained networks and randomly initialized networks and observe that pretraining does not improve the robustness. As a result, we observe that models trained from scratch for longer iterations achieve higher robustness than pretrained models.

\section{Conclusion}

This paper presented a comprehensive evaluation of instance segmentation models with respect to real-world image corruptions and images that exhibit domain gaps with the training datasets the models are learned from. 
We investigated different model ingredients in terms of their robustness, such as their architectural designs, network backbones, initializations,  normalization modules, and the effect of joint-task training. 
With these analyses, we provide insights into the robustness of state-of-the-art instance segmentation models.
These findings are essential for researchers when designing their models for applications where image corruptions are known in advance and when picking a trained off-the-shelf model to infer on the dataset at hand.

{\small
\bibliographystyle{ieee}
\bibliography{egbib}
}

\ifCLASSOPTIONcaptionsoff
  \newpage
\fi

\end{document}